\documentclass{article}

\usepackage[preprint]{neurips_2025}

\usepackage[utf8]{inputenc} 
\usepackage[T1]{fontenc}    
\usepackage{microtype}      

\usepackage{hyperref} 
\usepackage{url}            
\usepackage{orcidlink}      

\usepackage{xcolor}         
\usepackage{graphicx}       
\usepackage{tikz}           
\usepackage{epsfig}         
\usepackage{subfig}         
\usepackage{subcaption}     
\usepackage{caption}        
\usepackage{wrapfig}        
\usepackage{color}          
\usepackage{colortbl}       

\usepackage{amsmath}        
\usepackage{amssymb}        
\usepackage{amsfonts}       
\usepackage{amstext}        
\usepackage{amsthm}         
\usepackage{mathtools}      
\usepackage{bbm}            
\usepackage{bm}             
\usepackage{nicefrac}       
\usepackage{cases}          

\usepackage{booktabs}       
\usepackage{multirow}       
\usepackage{makecell}       

\usepackage{algorithm}      
\usepackage{algpseudocode}  
\usepackage{listings}       

\usepackage{ulem}           
\usepackage{soul}           
\usepackage{footnote}       
\usepackage{tablefootnote}  
\usepackage{comment}        

\definecolor{personColor}{RGB}{255,0,0}
\definecolor{laptopColor}{RGB}{0,0,255}
\definecolor{cellphoneColor}{RGB}{255,255,0}
\definecolor{pottedplantColor}{RGB}{0,255,0}
\definecolor{chairColor}{RGB}{255,0,255}

\usepackage[accsupp]{axessibility}  

\usepackage{pifont}         
\usepackage{wasysym}
\usepackage{enumitem}

\usepackage{longtable}
\usepackage{array}
\usepackage{booktabs} 
\usepackage{caption}  
\usepackage{ragged2e} 

\definecolor{lightblue}{RGB}{173,216,230}
\definecolor{lightgreen}{RGB}{144,238,144}
\definecolor{lightred}{RGB}{255,182,193}
\definecolor{lightyellow}{RGB}{255,255,224}
\definecolor{lightpurple}{RGB}{221,160,221}
\definecolor{lightgray}{RGB}{211,211,211}
\definecolor{lightorange}{RGB}{255,218,185}
\definecolor{lightpeach}{rgb}{1.0, 0.882, 0.788}
\definecolor{customblue}{rgb}{0.180, 0.400, 0.522}
\definecolor{lightcyan}{rgb}{0.8196, 0.9725, 0.9804}
\definecolor{sh_blue}{rgb}{0,0.60,0.93}
\definecolor{sh_red}{rgb}{0.8627, 0.3098, 0.3176}
\definecolor{highlight}{RGB}{255,255,0}
\definecolor{warning}{RGB}{255,99,71}
\definecolor{success}{RGB}{50,205,50}
\definecolor{info}{RGB}{30,144,255}
\definecolor{top1}{RGB}{255,179,179}
\definecolor{top2}{RGB}{255,217,179}
\definecolor{top3}{RGB}{255,255,179}
\definecolor{textblue}{RGB}{94,159,220} 
\definecolor{textgreen}{RGB}{59,125,35} 
\definecolor{textorange}{RGB}{192,80,21} 

\definecolor{tagred}{RGB}{196,15,15} 
\definecolor{tagblue}{RGB}{33,95,154} 

\definecolor{teaserblue}{RGB}{33,95,154} 

\definecolor{teasergree}{RGB}{57,158,163} 
\definecolor{teaserpurpe}{RGB}{105,111,173} 

\newcommand{\cmark}{\color{teasergree}\ding{51}}
\newcommand{\xmark}{\color{tagred}\ding{55}}

\newcommand{\param}[1]{\textcolor{teaserblue}{#1}}  

\definecolor{primary}{RGB}{70,130,180}
\definecolor{secondary}{RGB}{119,136,153}
\definecolor{accent}{RGB}{255,140,0}

\newcommand{\think}[1]{\textcolor{tagblue}{\texttt{#1}}}  
\newcommand{\answer}[1]{\textcolor{tagred}{\texttt{#1}}}  

\definecolor{customblue}{HTML}{E7EFFA}
\definecolor{custompink}{HTML}{F7E1ED}

\newcommand{\best}[1]{\colorbox{lightpeach!90}{#1}}  
\newcommand{\second}[1]{\colorbox{teasergree!20}{#1}}  

\newcommand*\colourcheck[1]{%
  \expandafter\newcommand\csname #1check\endcsname{\textcolor{#1}{\ding{52}}}%
}
\colourcheck{blue}
\colourcheck{green}
\colourcheck{red}

\title{JarvisArt: Liberating Human Artistic Creativity \\ via an Intelligent Photo Retouching Agent} 

\author{%
  Yunlong Lin$^{1\scalebox{0.75}{*}}$ \quad Zixu Lin$^{1\scalebox{0.75}{*}}$ \quad Kunjie Lin$^{1\scalebox{0.75}{*}}$ \quad Jinbin Bai$^{5}$ \quad Panwang Pan$^{4}$ \quad Chenxin Li$^{3}$ \\ \textbf{Haoyu Chen}$^{2}$ \quad \textbf{Zhongdao Wang}$^{6}$ \quad \textbf{Xinghao Ding}$^{1 \dag}$ \quad \textbf{Wenbo Li}$^{3 \scalebox{0.6}{$\clubsuit$}}$ \quad \textbf{Shuicheng Yan}$^{5 \dag}$\vspace{0.2cm} 
  \thanks{$*$ Equal Contributions. \quad $\clubsuit$ Project Leader \quad $\dagger$ Corresponding Authors.}
  \\
  $^{1}$ Key Laboratory of Multimedia Trusted Perception and Efficient Computing, \\
  Ministry of Education of China, Xiamen University, Xiamen, Fujian, China \\
  $^{2}$ The Hong Kong University of Science and Technology (Guangzhou)  \\
  $^{3}$ The Chinese University of Hong Kong \quad $^{4}$ Bytedance \\
  $^{5}$ National University of Singapore  \\
  $^{6}$ Tsinghua University \\
  \texttt{\small Project Page: \url{https://jarvisart.vercel.app/}} \\
  \vspace{-0.7cm}
}
\begin{document}

\maketitle
\begin{abstract}

Photo retouching has become integral to contemporary visual storytelling, enabling users to capture aesthetics and express creativity. While professional tools such as Adobe Lightroom offer powerful capabilities, they demand substantial expertise and manual effort. In contrast, existing AI-based solutions provide automation but often suffer from limited adjustability and poor generalization, failing to meet diverse and personalized editing needs. To bridge this gap, we introduce JarvisArt, a multi-modal large language model (MLLM)-driven agent that understands user intent, mimics the reasoning process of professional artists, and intelligently coordinates over 200 retouching tools within Lightroom. JarvisArt undergoes a two-stage training process: an initial Chain-of-Thought supervised fine-tuning to establish basic reasoning and tool-use skills, followed by Group Relative Policy Optimization for Retouching (GRPO-R) to further enhance its decision-making and tool proficiency. We also propose the Agent-to-Lightroom Protocol to facilitate seamless integration with Lightroom. To evaluate performance, we develop MMArt-Bench, a novel benchmark constructed from real-world user edits. JarvisArt demonstrates user-friendly interaction, superior generalization, and fine-grained control over both global and local adjustments, paving a new avenue for intelligent photo retouching. Notably, it outperforms GPT-4o with a \textbf{60\%} improvement in average pixel-level metrics on MMArt-Bench for content fidelity, while maintaining comparable instruction-following capabilities.

\end{abstract}

\section{Introduction}
Photo retouching is fundamental to modern photography, enabling users to manipulate exposure, color, contrast, and tone for expressive, high-quality images. Commercial tools such as Adobe Lightroom and PicsArt offer extensive manual controls but demand specialized expertise and significant time investment, creating barriers for non-experts. Existing automated methods—including zero- and first-order optimization~\cite{hansen2006cma,nishimura2018automatic,tseng2022neural,yu2021reconfigisp}, reinforcement learning~\cite{wu2024goal,kosugi2020unpaired,ke2022harmonizer}, and diffusion-based editing~\cite{zhang2023magicbrush,brooks2023instructpix2pix,xiao2024omnigen}—improve automation yet remain limited in stylistic diversity, fine-grained adjustment, and scene generalization. More recently, instruction-guided multimodal models such as GPT-4o~\cite{hurst2024gpt} and Gemini-2-Flash~\cite{team2023gemini} have enabled natural-language–driven editing but frequently compromise content fidelity, intricate attribute control, and high-resolution support.


LLM~\cite{guo2025deepseek,Cai2024Internlm2,yang2024qwen2,jarvisir2025}-powered agents have driven breakthroughs in autonomous task execution and problem solving, inspiring us to explore a novel photo-retouching paradigm: \textit{an intelligent, user-friendly artist agent that interprets the user’s intent and delivers professional-level edits}. To this end, we introduce JarvisArt, which (1) accurately parses visual inputs and natural-language instructions, (2) embeds professional retouching expertise to emulate an artist’s reasoning, (3) efficiently manages over 200 Lightroom operations, and (4) supports both global and local adjustments through an intuitive interface. All planning and tool invocations are fully transparent, allowing users to interactively refine the retouching workflow to suit their individual preferences.


\begin{figure*}[!t]
    \centering
\setlength{\abovecaptionskip}{0.1cm} 
    \includegraphics[width=1\linewidth]{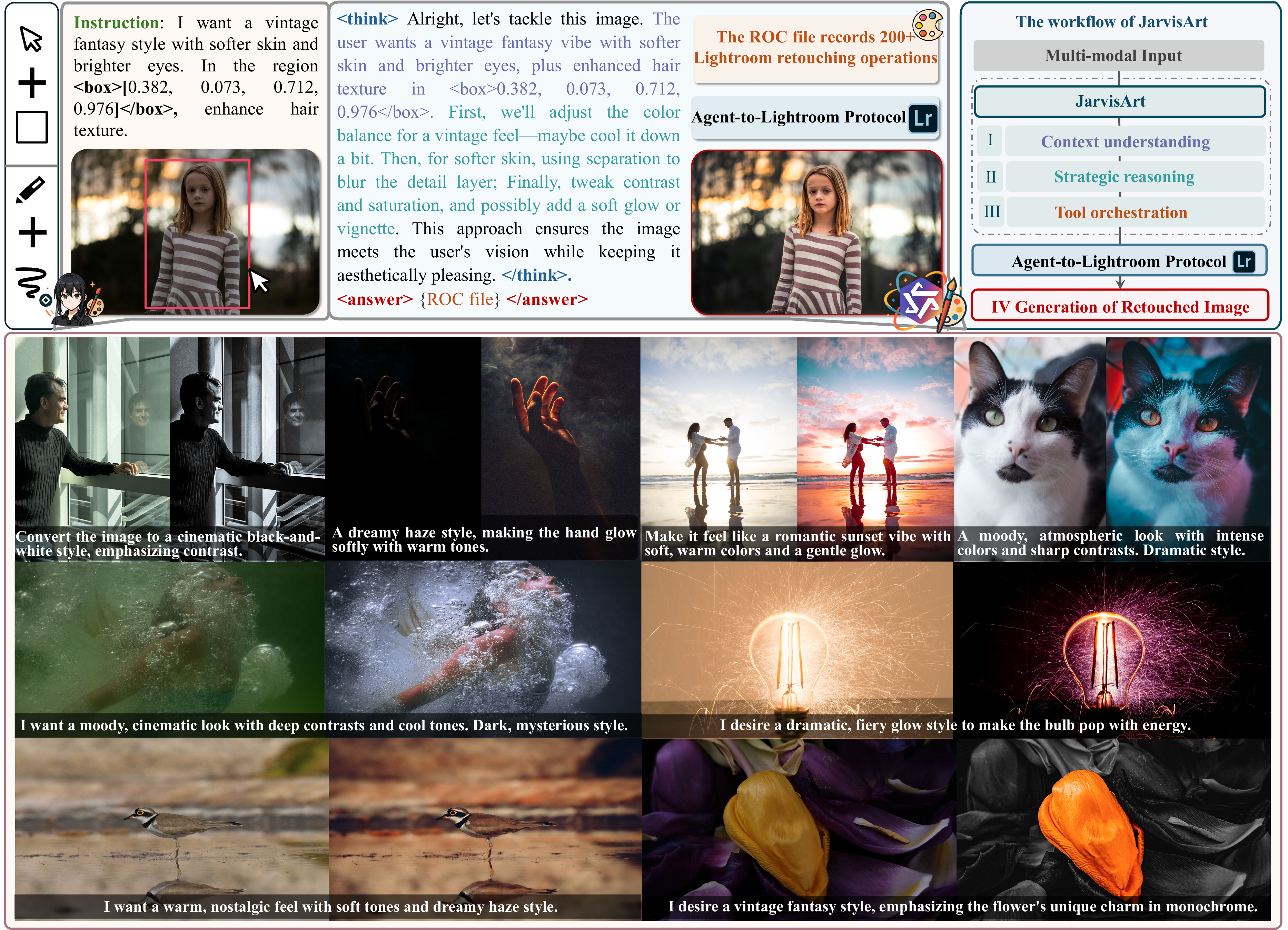}
    \caption{JarvisArt supports multi-granularity retouching goals, ranging from scene-level adjustments to region-specific refinements. Users can perform intuitive, free-form edits through natural inputs such as text prompts, bounding boxes, or brushstrokes. Furthermore, users can edit any-resolution images with JarvisArt. \textcolor{teaserpurpe}{Purple}: multi-modal context understanding. \textcolor{teasergree}{Green}: retouching strategy reasoning. \textcolor{textorange}{Orange}: decision-making in tool orchestration.}
    \label{fig:intro}
\end{figure*}

To translate this vision into practice, we must tackle three core challenges:the scarcity of high-quality data (source/target images, textual instructions, and editing configurations), the need for expert-level reasoning strategies, and the absence of a standardized Agent-to-Lightroom integration protocol. To overcome these, we first design a data-generation pipeline that yields the MMArt-55K dataset, comprising 5K standard and 50K Chain-of-Thought–enhanced multi-granularity samples. Next, we employ a two-stage post-training regime: (1) supervised fine-tuning (SFT) to instill a workflow of ``understanding $\rightarrow$ reasoning $\rightarrow$ decision-making (recording Lightroom operations into a ROC file)'', as illustrated in Figure~\ref{fig:intro}, and (2) Group Relative Policy Optimization for Retouching (GRPO-R) augmented with multi-dimensional tool-use rewards—namely, retouching-operation accuracy (evaluating both global and region-specific parameter prediction) and perceptual quality (assessing the visual fidelity of retouched outputs)—to refine decision-making and generalization. Finally, we introduce the Agent-to-Lightroom (A2L) protocol to enable seamless, automated Lightroom editing with bidirectional feedback. Consequently, JarvisArt deeply understands the intent of the user, generates diverse stylistic renditions, and seamlessly executes global and region-specific adjustments to produce visually compelling results (see Figure~\ref{fig:intro}).


Our contributions can be summarized as follows:
\begin{itemize}
\item We introduce JarvisArt, an intelligent artist agent powered by an MLLM and linked to over 200 Lightroom operations, capable of producing diverse, user-driven stylistic edits that surpass current automated methods and rival professional human retouchers.
\item We design a scalable data-synthesis pipeline to construct the MMArt dataset, comprising 5K standard instruction-based and 50K Chain-of-Thought–enhanced multi-granularity samples for detailed retouching tasks.
\item We develop a two-stage post-training regime: SFT followed by GRPO-R with tailored tool-use rewards to enhance the agent's reasoning, tool proficiency, and generalization.
\item We establish an Agent-to-Lightroom communication protocol that enables seamless collaboration between JarvisArt and Lightroom, facilitating fully automated editing workflows.
\end{itemize}

\section{Related Work}
\textbf{Photo Retouching.} 
Existing automated pipelines pipelines have been proposed to streamline manual retouching. Zeroth- and First-order optimizations~\cite{hansen2006cma,nishimura2018automatic,mosleh2020hardware,chen2017zoo,tseng2022neural,yu2021reconfigisp,tseng2019hyperparameter} were early attempts, but they are constrained by limited parameter prediction and reliance on pre-trained proxies. RL-based methods~\cite{wu2024goal,kosugi2020unpaired,ke2022harmonizer,hu2018exposure,ouyang2023rsfnet} attempt to mimic human workflows and offer some transparency but fail to capture artistic vision and lack deeper user interaction. Diffusion models~\cite{zhang2023magicbrush,brooks2023instructpix2pix,xiao2024omnigen} dominate high-fidelity image synthesis but rely on static prompts and lack multi-turn reasoning or flexible language alignment, limiting open-ended editing. Additionally, recent unified image editing models have achieved dual breakthroughs in comprehension and generation. Notable examples include closed-source models like GPT-4o~\cite{hurst2024gpt} and  Gemini-2-Flash~\cite{team2023gemini}, as well as open-source models such as Janus-Pro~\cite{chen2025janus}, UniTok~\cite{ma2025unitok}, QLIP~\cite{zhao2025qlip}, and VARGPT-v1.1~\cite{zhuang2025vargpt}. Despite these breakthroughs, three key limitations remain: (1) destructive editing by regenerating all pixels, compromising content preservation; (2) lack of interactive and interpretable local attribute control (e.g., softening or brightening skin); and (3) the absence of arbitrary-resolution editing due to generative model architectural constraints.
Conversely, our study presents an interactive and interpretable retouching paradigm that integrates multimodal understanding with expert-level editing tools for non-destructive photo retouching.  JarvisArt empowers users through a human-agent collaboration loop, enabling scene-level edits alongside precise region-specific tweaks-blending creative flexibility with the rigor of a professional workflow.

\textbf{Reinforcement Fine-Tuning.} 
Rule-based reinforcement fine-tuning, as demonstrated by OpenAI's o1~\cite{Jaech2024Openai} and Deepseek-R1~\cite{guo2025deepseek}, has shown impressive performance in tasks such as mathematical reasoning~\cite{Cai2024Internlm2, Jaech2024Openai, Yang2024Qwen25math, Ying2024Internlm}, and code generation~\cite{Hui2024Qwen25coder, Jiao2024Preference, Zhang2024Codedpo, Zhang2024o1coder}. Subsequent research has extended this approach to multimodal models, designing task-specific reward functions for visual perception tasks. These include correct class prediction in image classification~\cite{pan2025metaspatial, chen2025r1v, meng2025mmeureka}, Intersection-over-Union (IoU) metrics in image localization and detection~\cite{liu2025visualrft, huang2025visionr1, yang2025r1, shen2025vlm}, accurate click position prediction in GUI grounding tasks~\cite{lu2025uir1, xia2025gui}, and effective interaction with search engines to leverage up-to-date external information~\cite{jin2025search}. However, unlike these tasks with a single correct answer, our task involves tool-integrated retouching, which requires predicting multiple tools and their parameters. Designing effective reward signals to support learning in this setting remains an open and underexplored challenge. In this paper, we propose customized tool-use rewards, enabling JarvisArt to equip advanced artistic reasoning and tool invocation capabilities.

\textbf{LLM-Empowered Agent.} 
LLM-powered agents have revolutionized AI systems due to three key developments: 1) unprecedented reasoning capabilities of LLMs~\cite{guo2025deepseek, Cai2024Internlm2, yang2024qwen2}; 2) advancements in tool manipulation and environmental interaction~\cite{langchain,autogpt,hong2023metagpt,huang2024audiogpt,liu2024apigen} and 3) sophisticated memory architectures that support longitudinal experience accumulation~\cite{guo2024lightrag,zhang2024survey,xie2023openagents,yao2023react}. 
Despite these advancements, three fundamental limitations persist when applying LLM agents to professional photo retouching: 1) the absence of a domain-specific retouching knowledge base, which hinders accurate interpretation of user intent, 2) limited decision-making abilities in selecting suitable tools and determining precise parameter values, and 3) absence of standardized protocols to ensure compatibility with professional retouching software integrations.
To address these limitations, we propose JarvisArt, a powerful artistic agent that integrates three core capabilities: (1) professional retouching expertise for precise understanding of user instructions, (2) proficiency with commercial retouching tools in Lightroom, and (3) standardized communication protocols for seamless Lightroom integration.

\section{Method}
We begin by outlining the overall workflow of JarvisArt (Sec.~\ref{sec:workflow}). Next, we introduce a comprehensive data generation pipeline that constructs MMArt, a high-quality dataset comprising instruction and reasoning samples for agentic photo retouching tasks (Sec.~\ref{sec:mmart}). Finally, we investigate the core components of JarvisArt (Sec.~\ref{sec:framework}), including a two-stage post-training pipeline and the Agent-to-Lightroom (A2L) protocol, which allows seamless collaboration between JarvisArt and Lightroom.

\subsection{Overview}
\label{sec:workflow}
JarvisArt is an interactive, MLLM-based photo-retouching system that supports both scene-level and region-level edits. In addition to textual instructions, users can specify local areas via free-form brushstrokes or draggable bounding boxes. In Figure~\ref{fig:intro}, JarvisArt’s pipeline comprises three stages: 1) Multi-modal context understanding to parse user directives, image content, and regions of interest; 2) Strategic reasoning grounded in photographic principles to formulate a retouching plan; and 3) Tool orchestration to select appropriate Lightroom operations and parameters. These operations are executed automatically through the A2L protocol. Formally, JarvisArt implements a function:
\[
f(Q, I_{\text{src}}) \;\rightarrow\;\mathcal{T} = \{t_1, t_2, \ldots, t_n\},
\]
where \(Q\) is the user query, \(I_{\text{src}}\) the source image, and each \(t_i\) denotes a specific Lightroom edit (\textit{e.g.}, exposure \(\!+\!0.03\)). The final output is obtained by \(I_{\text{edit}} = g(I_{\text{src}}, \mathcal{T})\), with \(g(\cdot)\) representing Lightroom’s execution environment.

\begin{figure*}[!t]
    \centering
    \setlength{\abovecaptionskip}{0.1cm} 
    \setlength{\belowcaptionskip}{-0.4cm}
    \includegraphics[width=\linewidth]{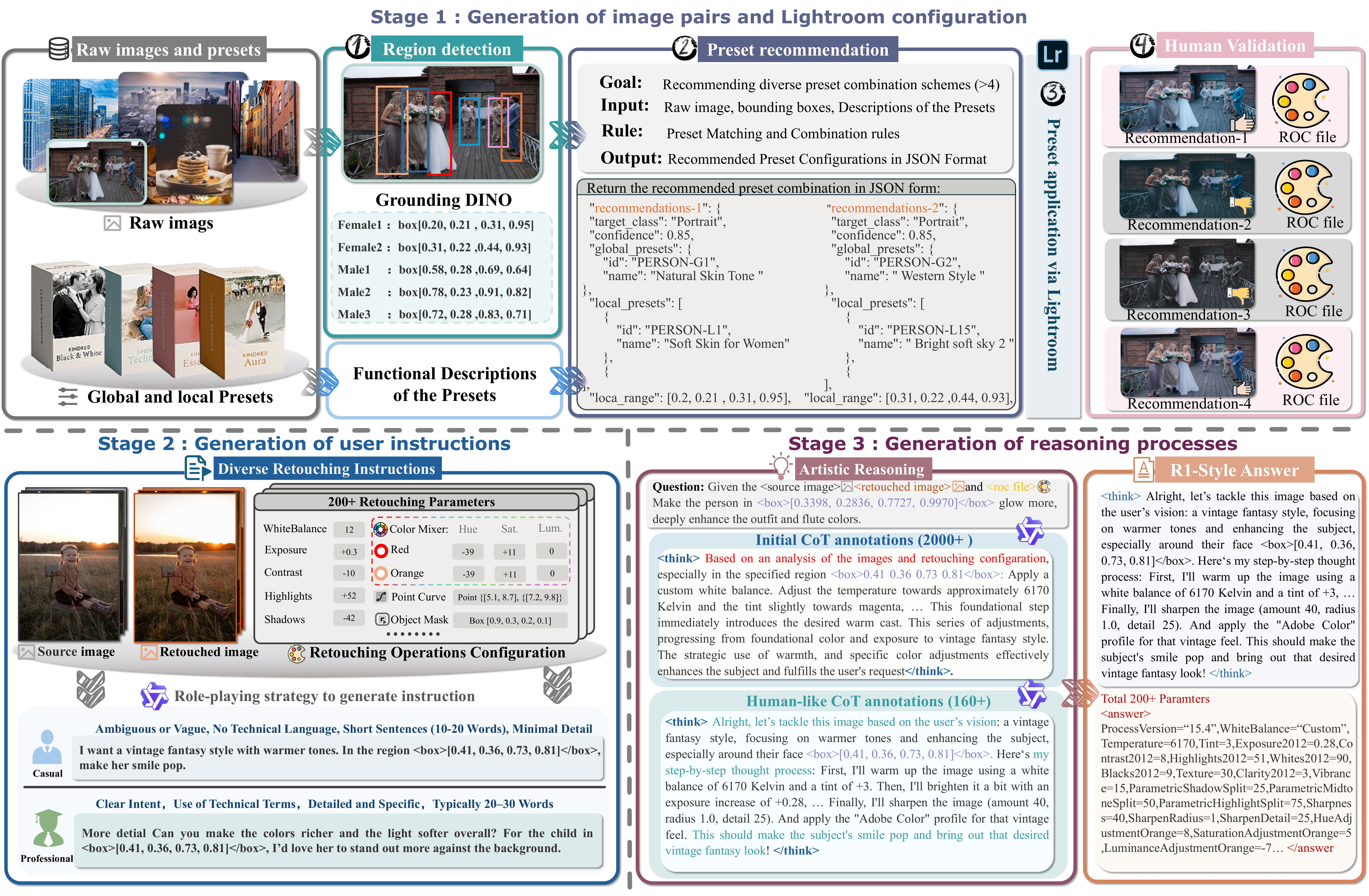}
    \caption{The data generation pipeline comprises three main stages: 1) Curation of diverse source–target examples covering varied scenes and styles with corresponding Lightroom configurations; 2) Generation of diverse user instrcutions that reflects different creative intents; 3) Production of Chain-of-Thought traces that simulate a human artist reasoning process.}
    \label{data_synthesis}
\end{figure*}

\subsection{Data Generation Pipeline}\label{sec:mmart}

We design a three-stage data-generation pipeline (Figure~\ref{data_synthesis}) to construct MMArt with explicit Chain-of-Thought (CoT) annotations. Each sample is a five-tuple \(\langle\mathit{I}_{\text{src}}, \mathit{I}_{\text{tgt}}, Q, \mathcal{C}, O\rangle\), where \(\mathit{I}_{\text{src}}\) and \(\mathit{I}_{\text{tgt}}\) are the before-/after-retouch images, \(Q\) the user’s instruction, \(\mathcal{C}\) the CoT reasoning wrapped in \think{<think>} tags, and \(O\) the retouching operation configuration (ROC) file of tool invocations and parameters within \answer{<answer>} tags. The pipeline proceeds as follows: 1) Curation of diverse source–target examples covering varied scenes and styles, and the corresponding Lightroom configurations; 2) Generation of natural-language instructions that reflect user intents; 3) Production of step-by-step reasoning traces. Further statistics and examples of MMArt can be found in Appendix~\ref{app:MMart_dataset}.

\textbf{Stage I: Generation of image pairs and Lightroom configuration.} We source raw images from PPR10K~\cite{liang2021ppr10k}, the Adobe Lightroom community, and licensed open-source collections, then curate a diverse library of global and local artistic presets. Leveraging Qwen2.5-VL-72B~\cite{yang2024qwen2} for multimodal role-playing and Grounding DINO~\cite{liu2024grounding} for precise region localization, we simulate expert-level edits in four steps: 1) \textit{Region detection}, in which Grounding DINO~\cite{liu2024grounding} identifies regions of interest (confidence > 0.8); 2) \textit{Preset recommendation}, where Qwen2.5-VL-72B~\cite{yang2024qwen2} proposes global and local presets based on image aesthetics; 3) \textit{Preset application}, applying each recommendation in Lightroom to generate five candidate retouched images; and 4) \textit{Human-in-the-loop validation}, selecting the most artistically pleasing outputs. Each finalized sample comprises \(\langle I_{\text{src}}, I_{\text{tgt}}, O\rangle\), denoting the source image, the retouched image, and the detailed record of Lightroom operations. The role-playing prompts are detailed in Appendix~\ref{app:recommed_temple}.

\textbf{Stage II: Generation of user instructions.}
To simulate diverse editing intents, we employ Qwen2.5-VL-72B~\cite{yang2024qwen2} with a role-playing prompt (Appendix~\ref{app:recommed_temple}) to translate each \(\langle I_{\text{src}}, I_{\text{tgt}}, O\rangle\) triplet into both scene-level and region-level instructions \(Q\). We generate descriptions for two user types—casual users and professional editors with advanced aesthetic sensibilities, ensuring coverage of simple global edits as well as precise, localized adjustments.

\textbf{Stage III: Generation of reasoning processes.} For each sample quadruple \(\langle I_{\text{src}}, I_{\text{tgt}}, Q, O\rangle\), we first apply QVQ-max’s~\cite{yang2024qwen2} advanced visual reasoning to generate initial CoT annotations. To remove redundancy and enforce human-like coherence, we subsequently refine these traces using Qwen2.5-VL-72B~\cite{yang2024qwen2} through iterative multimodal prompts, producing concise, context-rich reasoning processes \(\mathcal{C}\). Full prompt templates are provided in Appendix~\ref{app:recommed_temple}.

\subsection{JarvisArt Framework}\label{sec:framework}

\subsubsection{CoT Supervised Fine-tuning}\label{SFT}
Drawing on Deepseek-R1~\cite{guo2025deepseek}, we initialize JarvisArt via supervised fine-tuning on CoT annotations to bootstrap its subsequent reinforcement learning. This phase 1) enforces a consistent, structured output format, 2) instills foundational reasoning skills spanning user-intent interpretation and aesthetic judgment, and 3) establishes preliminary proficiency in selecting Lightroom tools and configuring their parameters.




\begin{figure*}[!t]
    \setlength{\abovecaptionskip}{0.1cm} 
    \setlength{\belowcaptionskip}{-0.6cm}
    \centering
    \includegraphics[width=\linewidth]{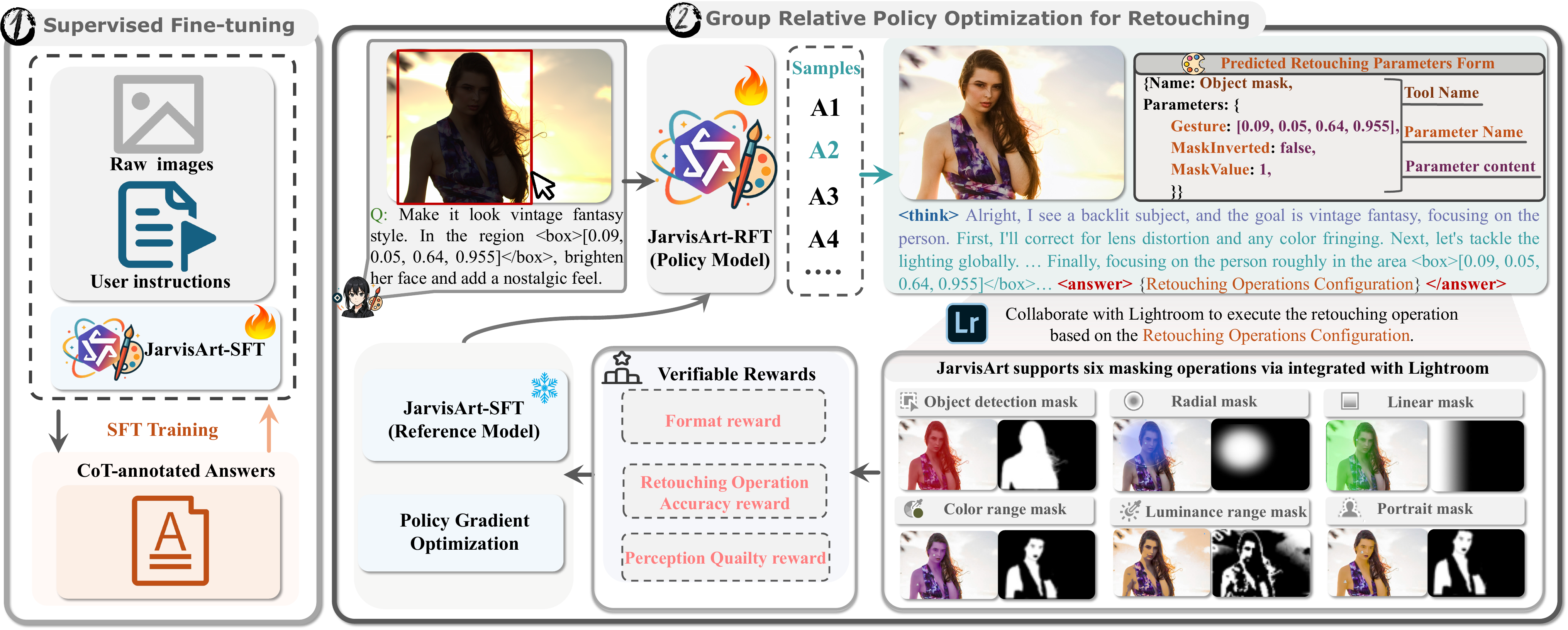}
    \caption{Overview of the two-stag post-training framework. Initially, JarvisArt undergoes supervised fine-tuning (SFT) on CoT-annotated data to develop foundational artistic reasoning and tool-use skills. Following this, we apply the Group Relative Policy Optimization for Retouching (GRPO-R) algorithm to further enhance the JarvisArt's reasoning, tool proficiency, and generalization.
    }
    \label{framework_fig}
\end{figure*}


\subsubsection{Reasoning-oriented Reinforcement Learning}\label{RFT}
Building on the SFT-initialized model, as shown in Figure~\ref{framework_fig}, we apply group relative policy optimization for retouching(GRPO-R)~\cite{shao2024deepseekmath} (Appendix~\ref{app:GRPO_detials}) to further refine JarvisArt’s artistic reasoning and tool-use proficiency. GRPO-R trains the agent with three interpretable, task-specific rewards: a format reward \(R_f\) that enforces structured output, a retouching operation accuracy reward \(R_{roa}\) that measures the correctness of selected tools and their parameter settings, and a perceptual quality reward \(R_{pq}\) that assesses the visual fidelity of the retouched image. The overall objective is thus $R = R_{f} + R_{roa} + R_{pq} \in [0,3]$.

\textbf{Format reward.} Following prior work~\cite{guo2025deepseek,yang2025r1,shen2025vlm,xia2025gui}, we include a format reward \(R_{f} \in [0,1]\) to enforce structured outputs: reasoning must appear within \think{<think>} tags and tool invocations within \answer{<answer>} tags, ensuring consistent and reliable parsing.

\textbf{Retouching operation accuracy reward.} Inspired by existing explorations of reward designs~\cite{qian2025toolrl,jin2025search,lu2025uir1, xia2025gui} in the fields of GUI and web searching.
We consider over 200 retouching tools in Lightroom, containing both global adjustments—such as exposure, highlights, and tone curve—and local refinements using six types of masks: 1) linear masks for directional gradients, 2) radial masks for circular or elliptical regions, 3) object masks for isolating subjects (\textit{e.g.}, people or objects), 4) color masks for hue-specific adjustments, 5) luminance masks for brightness-based selections, and 6) portrait masks for fine-tuning facial features such as skin and eyes. Further details are provided in Appendix~\ref{app:tools_detials}. To assess the accuracy of predicted tools and their parameters, $T^{pre} = \{T^{pre}_1, ..., T^{pre}_M\}$, against the ground truth $T^{tgt} = \{T^{tgt}_1, ..., T^{tgt}_N\}$, we define the ROA reward based on three evaluation criteria:

\ding{202} \textit{Tool name matching:}
\begin{equation}
r_{\text{name}} = \frac{|N_{T^{pre}} \cap N_{T^{tgt}}|}{|N_{T^{pre}} \cup N_{T^{tgt}}|} \in [0, 1] \,,
\end{equation}
where $N_{T^{pre}}$ and $N_{T^{tgt}}$ are the sets of tool names in the predicted and target sequences, respectively.

\ding{203} \textit{Parameter name matching:}
\begin{equation}
r_{\text{param}} = \sum_{T^{tgt}_j \in T^{tgt}} \sum_{T^{pre}_i \in T^{pre}} \frac{|\text{keys}(T^{pre}_i) \cap \text{keys}(T^{tgt}_j)|}{|\text{keys}(T^{pre}_i) \cup \text{keys}(T^{tgt}_j)|} \in [0, |T^{tgt}|] \,,
\end{equation}
where $\text{keys}(\cdot)$ denotes the set of parameter names associated with a predicted or ground-truth tool. It is noted that an overlap in parameter names occurs only when the predicted and ground-truth tool names match.

\ding{204} \textit{Parameter value matching:}
\begin{equation}
r_{\mathrm{value}}=\sum_{T^{tgt}_j\in T^{tgt}} \sum_{T^{pre}_i\in T^{pre}} \sum_{k\in \mathrm{keys(T^{tgt}_j)}} {S_k\left( T^{pre}_{i}[k],T^{tgt}_{j}[k] \right) } \,\,\in [0,\sum_{T^{tgt}_j\in T^{tgt}}{|}\mathrm{keys(}T^{tgt}_j)|] \,,
\end{equation}
where $S_k\left( \cdot \right) \in [0,1]$ quantifies the correspondence between predicted and ground-truth parameter values, with a value of $1$ indicating an exact match. Specifically, if the key $k$ is absent in $T^{pre}_{i}$, then $T^{pre}_{i}[k]$ is undefined and $S_k=0$. The computation of $S_k$ depends on the parameter type: scalar differences for standard numerical values, intersection-over-union (IoU) for object masks, endpoint distance for linear masks, geometric similarity for radial masks, color distance between sampled points for color masks, luminance range differences for luminance masks, and category-specific criteria for portrait masks. Refer to Appendix~\ref{app:rewards_calculation} for further details.
Finally, the retouching operation accuracy reward is computed by measuring the matching degree between $T^{pre}$ and $T^{tgt}$:
\begin{equation}
    R_{\mathrm{roa}}=\frac{1}{3}\left( r_{\mathrm{name}}+\frac{r_{\mathrm{param}}}{|T^{tgt}|}+\frac{r_{\mathrm{value}}}{\sum_{T_{j}^{tgt}\in T^{tgt}}{|}\mathrm{keys(}T_{j}^{tgt})|} \right) \in [0,1]\,.
\end{equation}


\textbf{Perception quality reward}. While parameter-based rewards offer critical guidance, they may not fully capture the perceptual quality of the final image, as different parameter settings can produce visually similar results. To address this limitation, we introduce the PQ reward, which evaluates two key aspects: 1) global tone consistency via color distribution matching, and 2) pixel-wise fidelity. The reward is defined as:
\begin{equation}
R_{pq} = \gamma \cdot \text{CD}(I_{\text{edit}},I_{\text{tgt}}) + (1-\gamma) \cdot 
\text{L}(I_{\text{edit}},I_{\text{tgt}}) \in [0,1]\,,
\end{equation}
where \( I_{\text{edit}} \) is the retouched image and \( I_{\text{tgt}} \) is the target image. $\text{CD}(\cdot)$ measures color distribution similarity in CIELAB space~\cite{zhang1996spatial} and $\text{L}(\cdot)$ denotes the pixel-wise distance. Both metrics are normalized to the range [0, 1], with higher values indicating better similarity. 
The weighting factor is empirically set to $\gamma=0.4$ to balance both terms.

\begin{wrapfigure}{r}{0.4\textwidth}
    \vspace{-1.2cm}
    \centering
    \setlength{\belowcaptionskip}{-0.3cm}
    \includegraphics[width=\linewidth]{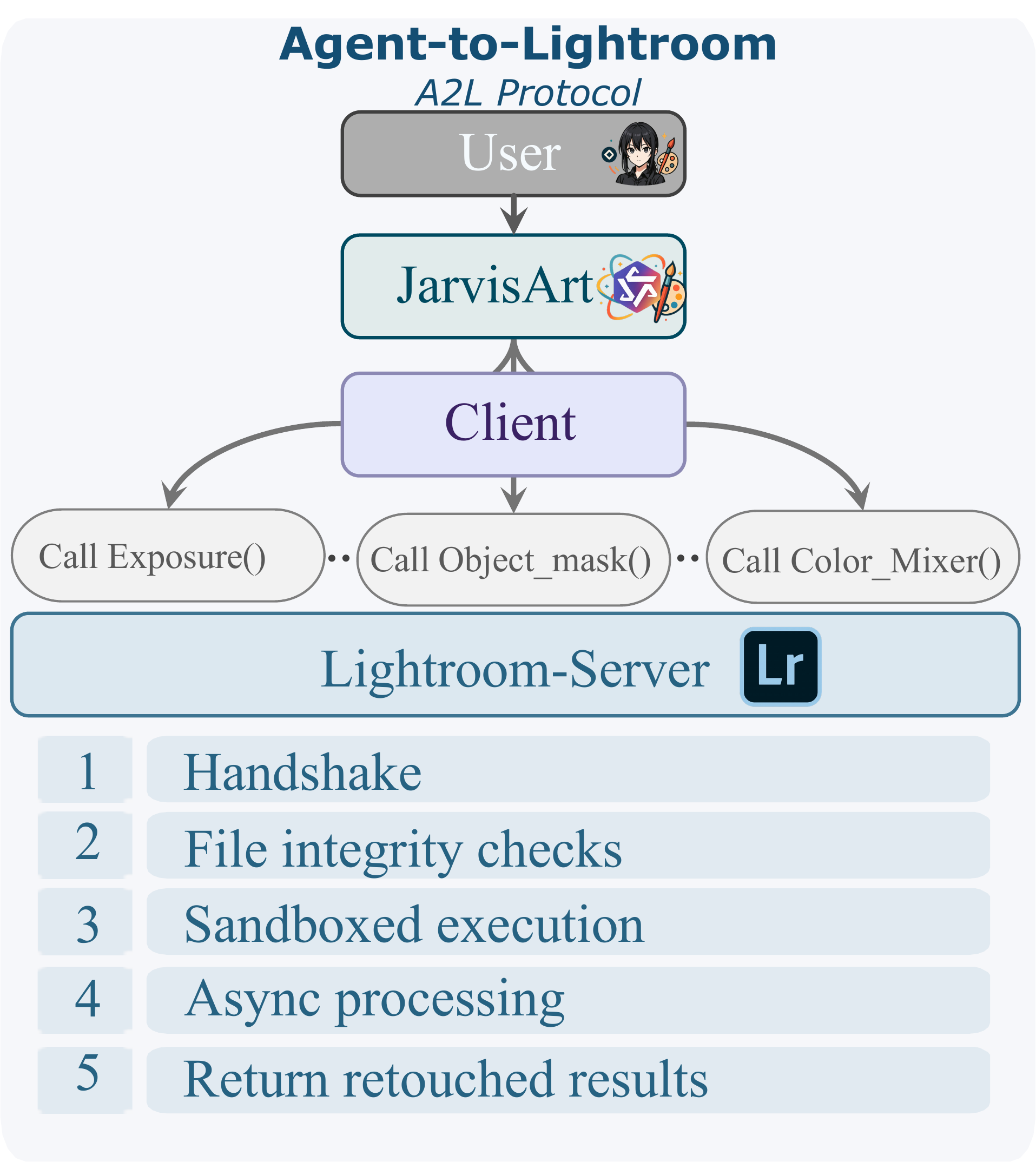}
    \caption{Agent-to-Lightroom protocol.}
    \label{a2l_fig}
\end{wrapfigure}
\subsubsection{Agent-to-Lightroom Protocol}
Figure~\ref{a2l_fig} presents the Agent-to-Lightroom (A2L) protocol, a standardized client-server interface that integrates JarvisArt with Lightroom. The workflow comprises five stages: 1) handshake, 2) file verification, 3) sandboxed execution, 4) async processing, and 5) result return. A2L features dual-transport communication, a structured message format, and resource management. Messages use bar-delimited commands for processing, status, and error handling, enhancing clarity and efficiency. It manages source images and retouching operation configuration (ROC) files, supporting ROC-to-Lua translation, and integrity checks. The source image can by directly retouched by Lua file in Lightroom. The Lua file can be directly applied in Lightroom to retouch the source image. Additional details are provided in the supplementary materials.

\section{Experiment}
\subsection{Experimental Setup}
\noindent\textbf{Implementation details.}
We adopt Qwen2.5-VL-7B-Instruct \cite{bai2025qwen2} as the base model for JarvisArt. The CoT supervised fine-tuning phase is performed on 50K CoT-annotated instances from MMArt, with a batch size of 2, a learning rate of 1e-5, and training for 2 epochs using the Llama-Factory framework \cite{zheng2024llamafactory} on 8 A100 (80G) GPUs. The reinforcement learning phase, employing the GRPO-R algorithm, is conducted on 5K standard instruction samples from MMArt, using the veRL framework \cite{sheng2024hybridflow}. For each training step, we sample a batch of 2, a learning rate of 1e-6, and generate 4 responses per query, training for 2 epochs on 16 A100 (80G) GPUs. 







\noindent\textbf{MMArt-Bench.}
To provide a comprehensive evaluation of JarvisArt's performance, we introduce the MMArt-Bench, which is sampled from the MMArt dataset. It includes four main scenarios: portrait, landscape, street scenes, and still life, with 50 instances per category, totaling 200 instances. Each primary category contains multiple subcategories (Appendix~\ref{app:MMart_Statistics}).
For region-level evaluation, we utilize a portrait subset comprising 50 human-centered images with mask annotations.

\noindent\textbf{Evaluation metrics.}
Following previous works~\cite{zhang2023magicbrush,ku2023viescore}, six assessment metrics are use for evaluation: L1, L2, SC, PQ, and O. L1 and L2 to measure the average pixel-level absolute difference between the retouched image and reference image. SC evaluates the alignment between the instruction text and the image (0–10 scale). PQ evaluates contextual coherence and artifact presence (0–10 scale). The overall score O is calculated as $\text{O} = \sqrt{\text{SC} \times \text{PQ}}$. For region-specific evaluation, we apply these six metrics to a specified mask region. Further details are provided in Appendix~\ref{app:metric_calculate}.


\begin{table*}[t!]
    \centering
    \setlength{\abovecaptionskip}{0.1cm} 
    \setlength{\belowcaptionskip}{-0.8cm}
    \caption{Quantitative evaluation on MMArt-Bench. We highlight the \colorbox{lightpeach!90}{best} and \colorbox{teasergree!20}{second-best} instruction-based results. SC, PQ, and O refer to the metrics evaluated by Gemini-2-Flash. The $RC$ means the metric calculated on specific mask region. }\label{main_table}
    \scalebox{0.8}{
    \setlength\tabcolsep{2pt}
    \renewcommand\arraystretch{1}
    \begin{tabular}{lcccccccccccc}
    \toprule
    ~ & ~ & \multicolumn{5}{c}{Scene-level} & \multicolumn{5}{c}{Region-level} \\ 
    \cmidrule(lr){3-7} \cmidrule(lr){8-12} 
    \multirow{-2}{*}{Method} & \multirow{-2}{*}{Instruction} & $\text{L1}_{\times 10^{2}}$ $\downarrow$ & $\text{L2}_{\times 10^{3}}$ $\downarrow$ & SC $\uparrow$ & PQ $\uparrow$  & O $\uparrow$ & $\mathrm{L}1_{\times 10^{2}}^{RC}$ $\downarrow$ & $\mathrm{L}2_{\times 10^{3}}^{RC}$ $\downarrow$  & $\text{SC}^{RC}$ $\uparrow$ & $\text{PQ}^{RC}$ $\uparrow$ & $\text{O}^{RC}$ $\uparrow$ \\
    \midrule
    RSFNet~\cite{ouyang2023rsfnet}    & \xmark  & 11.61  & 26.38 & - & - & - & 8.80 & 13.69 & - & - & - &  \\
    3DLUT~\cite{zeng2020learning}     & \xmark & 11.50  & 25.99 & - & - & - & 8.33 & 12.39 & - & - & - &  \\
    \midrule
    InstructPix2Pix~\cite{brooks2023instructpix2pix}  & \cmark  & \second{15.67}  & \second{47.51} & 6.54 & 7.79 & 7.10 & 12.62 & 33.39 & 4.70 & 5.36 & 4.91 \\
    MagicBrush~\cite{zhang2023magicbrush} & \cmark & 18.39  & 65.25 & 3.93 & 4.09 & 3.85 & \second{12.37} & 32.81 & 3.04 & 3.41 & 3.13  \\
    OmniGen~\cite{xiao2024omnigen}  & \cmark  & 28.49  & 133.45 & 4.25 & 4.42 & 4.13 & 25.16 & 109.10 & 6.17 & 7.56 & 6.72  \\
    VARGPT-v1.1~\cite{zhuang2025vargpt}  & \cmark  & 27.05  & 126.47 & 1.83 & 1.38 & 1.48 & 23.71 & 107.32 & 1.38 & 1.15 & 1.08  \\
    Step1X-Edit~\cite{liu2025step1x}  & \cmark  & 24.28  & 105.91 & 7.52 & 8.67 & 8.01 & 15.43 & 45.85 & \second{8.32} & 9.04 & 8.66  \\
    \midrule
    Gemini-2-Flash~\cite{team2023gemini} & \cmark  & 23.07  & 90.99 & \second{7.62} & 8.78 & 8.08 & 16.52 & 52.88 & 8.04 & 9.25 & 8.61 \\
    GPT-4o~\cite{hurst2024gpt}  & \cmark  & 22.84  & 92.23 & \best{8.73} & \second{9.66} & \best{9.18} & 15.71 & 47.87 & \best{8.59} & \best{9.48} & \best{9.03}  \\
    \midrule

   \textbf{JarvisArt}& \cmark & \best{12.44} & \best{30.56} &  7.53 &  \best{9.82} & \second{8.52} & \best{7.63} & \best{12.14} & 8.08 & \second{9.39} &  \second{8.69} \\
    \bottomrule
    \end{tabular}}
\end{table*}
\begin{figure*}[!t]
    \vspace{-0.5cm}
    \setlength{\abovecaptionskip}{0.1cm} 
    \setlength{\belowcaptionskip}{-0.5cm}
    \centering
    \includegraphics[width=0.95\linewidth]{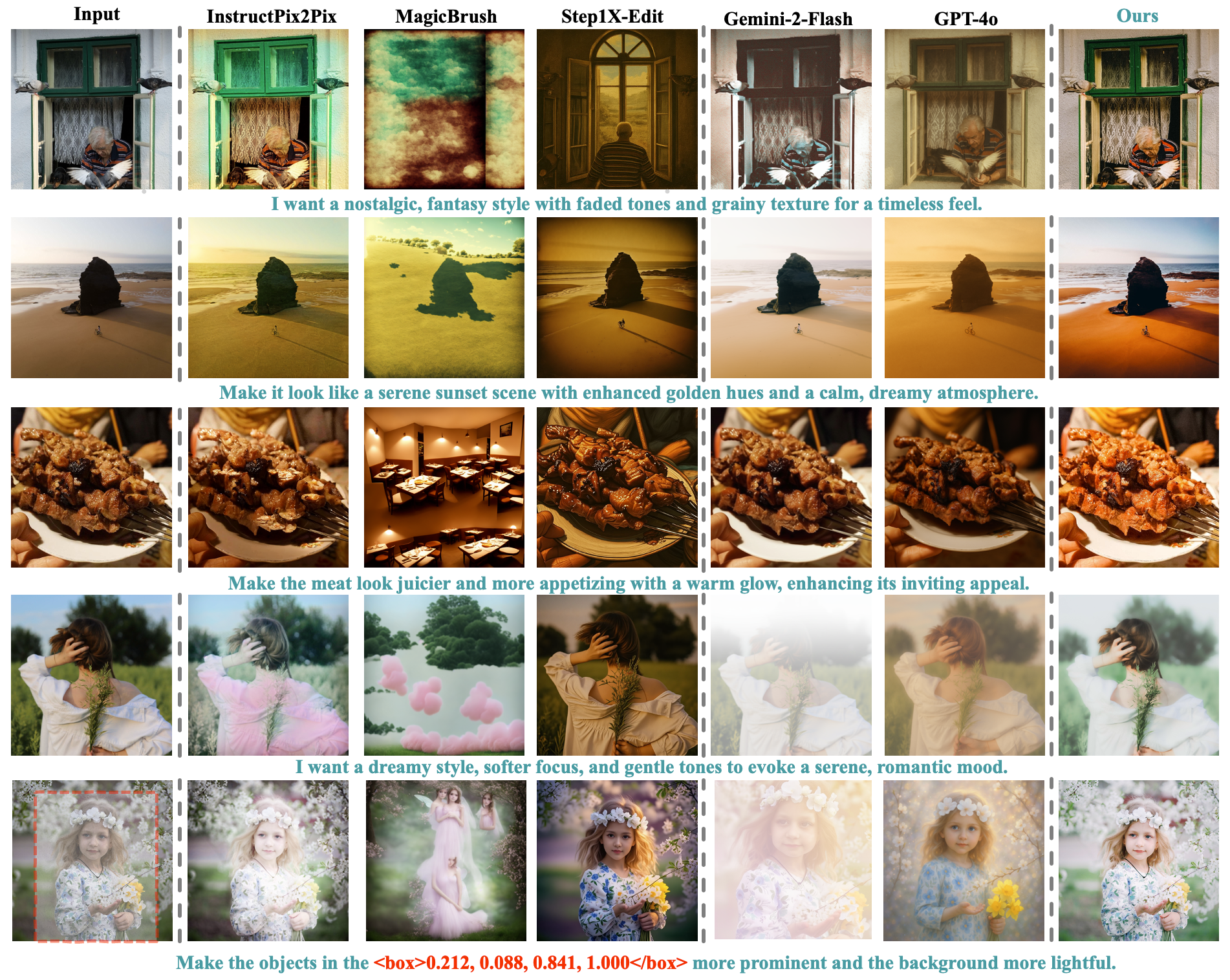}
    \caption{Visual comparison of different methods on MMArt-Bench.
    }
    \label{fig:visual_results}
\end{figure*}

\noindent\textbf{Baselines.} 
For a fair comparison, we evaluate JarvisArt against leading open-source photo retouching methods, including 3DLUT~\cite{zeng2020learning} and RSFNet~\cite{ouyang2023rsfnet}, as well as instruction-driven editing models such as InstructPix2Pix~\cite{brooks2023instructpix2pix}, MagicBrush~\cite{zhang2023magicbrush}, OmniGen~\cite{xiao2024omnigen}, VARGPT-v1.1~\cite{zhuang2025vargpt} and Step1X-Edit~\cite{liu2025step1x}. Proprietary solutions such as GPT-4o\footnote{The results are obtained based on ChatGPT APP in May 2025.}~\cite{hurst2024gpt} and Gemini-2- Flash~\cite{team2023gemini}\footnote{The results are obtained based on Gemini API in May 2025.} are also included for comparison. Notably, all test images are cropped to a $512 \times 512$ resolution, as some baselines are incapable of processing high-resolution or arbitrarily sized inputs.



\subsection{Experimental Results}
\subsubsection{Evaluation on MMArt-Bench}
As shown in Table~\ref{main_table}, JarvisArt outperforms most open-source instruction-based baselines, achieving state-of-the-art performance across all 10 evaluation metrics. Compared to closed-source models such as GPT-4o~\cite{Jaech2024Openai} and Gemini-2-Flash~\cite{team2023gemini}, JarvisArt achieves superior content preservation—for instance, an $\text{L1}_{\times 10^{2}}$ score of 12.44, which is 45.6\% lower (and thus better) than GPT-4o’s score of 22.84. JarvisArt also demonstrates competitive instruction-following capability (O = 8.52), closely matching GPT-4o (O = 9.18) and outperforming Gemini-2 Flash (O = 8.08). Notably, in the local editing setting—where content fidelity is especially critical—the advantage of our method over GPT-4o and Gemini-2-Flash is significantly amplified.
As illustrated in Figure~\ref{fig:visual_results}, especially in portrait scenarios, competing methods often exhibit noticeable uncanny valley effects, producing significant visual artifacts that diverge from users’ creative intent. In contrast, JarvisArt mitigates these issues through its Lightroom-integrated workflow, enabling high-quality, non-destructive editing. More results in Appendix~\ref{app:more_results}.

\begin{figure*}[!t]
    \centering
    \setlength{\abovecaptionskip}{-0.1cm} 
    \includegraphics[width=\linewidth]{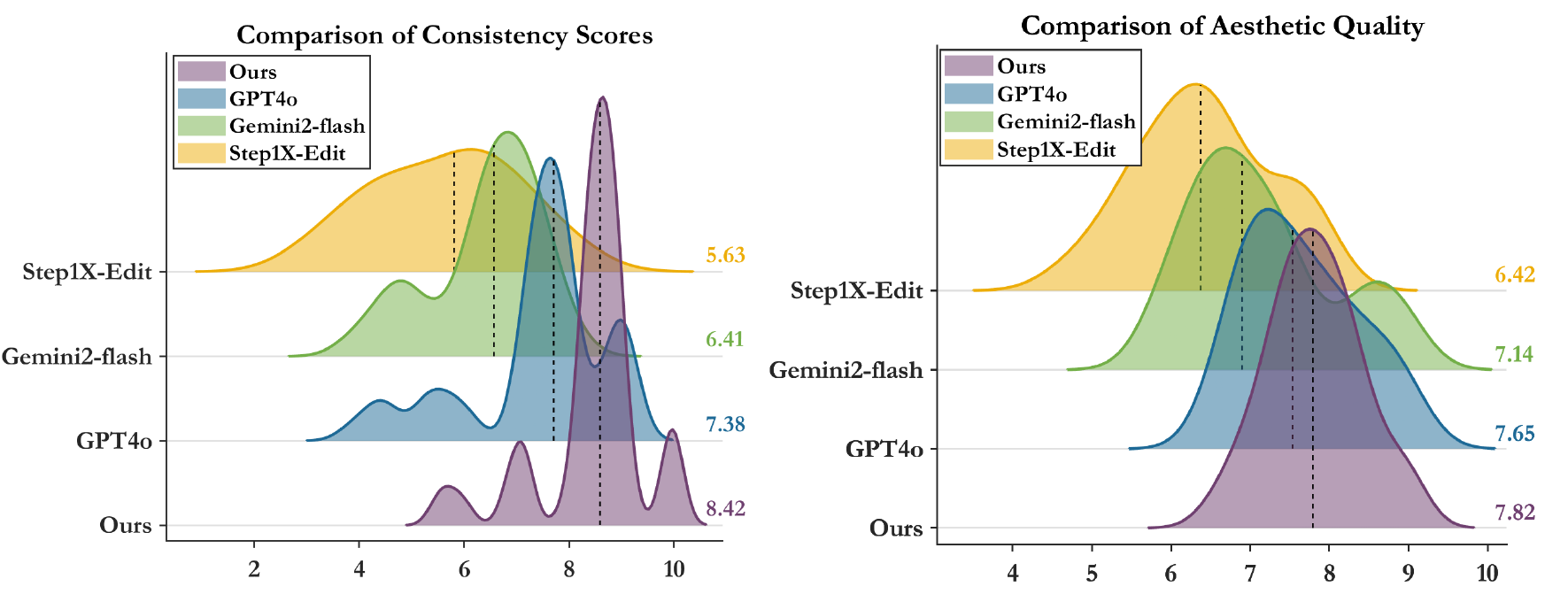}
    \caption{User preference study.}
    \label{user_study_fig}
\end{figure*}

\begin{figure*}[!t]
  \setlength{\belowcaptionskip}{-0.5cm}
  \centering
  \includegraphics[width=0.95\linewidth]{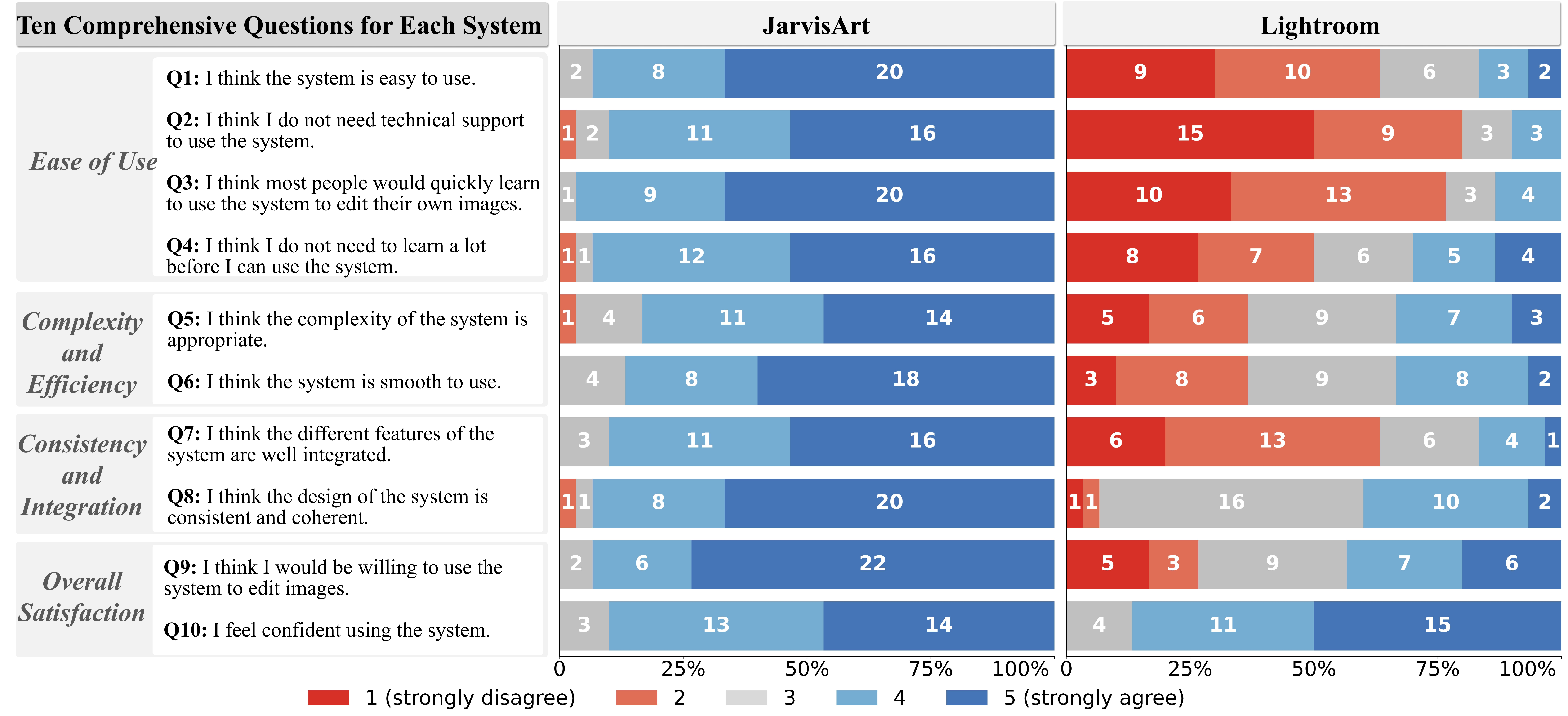}
  \caption{Questionnaire results and user ratings comparing JarvisArt with the commercial Adobe Lightroom system. Ratings are on a 5-point Likert scale (1=strongly disagree, 5=strongly agree).
  }
  \label{fig:user_study}
\end{figure*}

\subsubsection{User Preference Study}
Evaluating instruction-driven photo retouching remains inherently subjective, as even expert evaluators often disagree on the "optimal" outcome. To quantify preferences, we conducted a user preference study on the MMArt-Bench, recruiting 80 participants to evaluate four advanced algorithms: Step1X-Edit~\cite{liu2025step1x}, Gemini-2-Flash~\cite{team2023gemini}, GPT-4o~\cite{hurst2024gpt}, and JarvisArt. Evaluations focus on two criteria: (1) image consistency (preservation of source image content) and (2) aesthetic quality (visual appeal of retouched results). A five-point ordinal scale (worst = 2, poor = 4, fair = 6, good = 8, and excellent = 10) for quantitative metrics. Results in Figure~\ref{user_study_fig} show JarvisArt achieves best subjective quality, producing edits favored by users. 

To evaluate the effectiveness and usability of JarvisArt, we recruited 30 participants from diverse backgrounds, including postgraduate students, artists, and computer vision researchers. All participants had prior experience with image editing, covering a broad spectrum of skill levels to ensure a realistic representation of user proficiency. To mitigate learning effects, participants were randomly assigned to two groups: Group A used JarvisArt before Lightroom, while Group B used Lightroom first and then JarvisArt. Each participant completed a comprehensive evaluation consisting of 10 questions per system, covering four key categories: \textit{Complexity and Efficiency}, \textit{Consistency and Integration}, \textit{Ease of Use}, and \textit{Overall Satisfaction}. Detailed evaluation results are shown in Figure~\ref{fig:user_study}. The main findings are summarized as follows:
\begin{itemize}
    \item \textbf{Ease of use:} All participants rated JarvisArt as easy to use (Q1, score $\geq$ 3), with 66.7\% awarding the highest score of 5. For independent operation and learning speed (Q2–Q4), over 90\% of users rated JarvisArt 4 or 5, indicating that most users could quickly and independently learn to use the system with minimal need for technical support.
    \item \textbf{Complexity and efficiency:} Regarding system complexity (Q5), more than 96.67\% of participants (score $\geq$ 3) considered JarvisArt appropriately complex, in contrast to Lightroom, which was frequently perceived as overly complicated. Additionally, 86.67\% of users rated JarvisArt 4 or 5 for smoothness of use (Q6), suggesting that our design effectively reduced cognitive load and facilitated efficient task completion.

    \item \textbf{Consistency and integration:} For feature integration (Q7), 90\% of users rated JarvisArt 4 or 5, and for system consistency (Q8), 93.3\% did so, both significantly higher than Lightroom (16.67\% and 40\%, respectively). These results indicate a more cohesive and intuitive user experience with JarvisArt.
    \item \textbf{Overall satisfaction:} In terms of willingness to use in the future (Q9), 93.33\% of users rated JarvisArt 4 or 5, and for confidence in use (Q10), 90\% also gave high scores, both outperforming Lightroom (43.3\% and 86.7\%, respectively). This demonstrates strong user satisfaction and acceptance of JarvisArt.

\end{itemize}

\subsubsection{Visualization of Reward Trends for GRPO-R} 
\begin{figure*}[!t]
    \setlength{\abovecaptionskip}{-0.1cm} 
    \setlength{\belowcaptionskip}{-0.6cm}
    \centering
    \includegraphics[width=\linewidth]{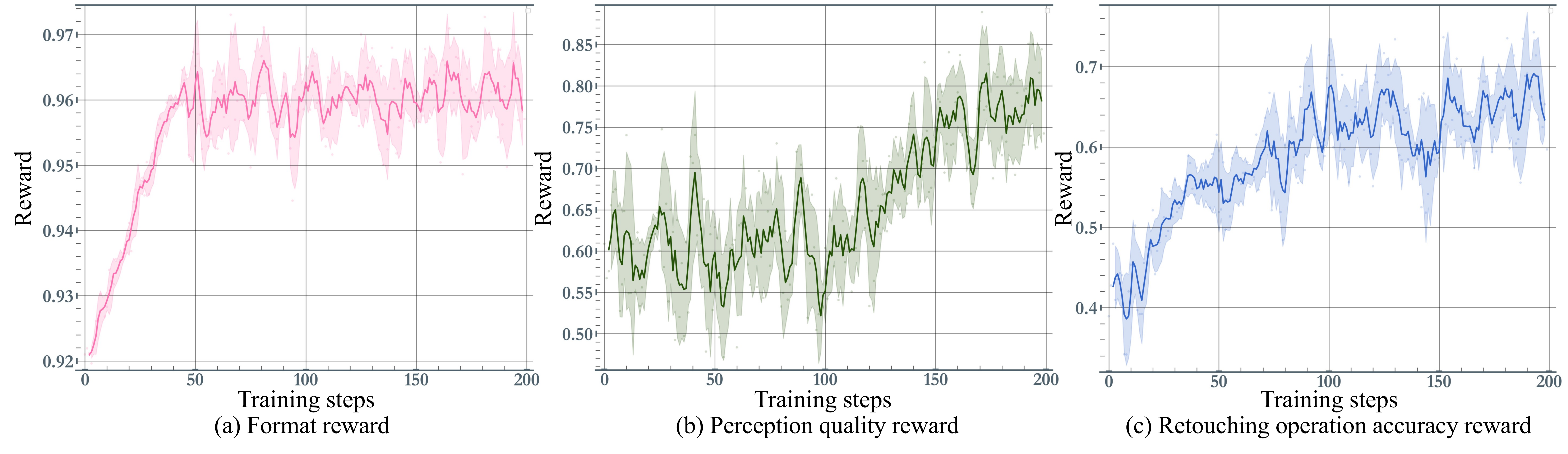}
    \caption{Visualization of the reward trends across training steps of for JarvisArt.}
    \label{fig:ablation}
\end{figure*}

Figure~\ref{fig:ablation} shows additional visualizations of GRPO-R training. The format reward converges quickly early on. While the PQ reward initially fluctuates and grows gradually, the ROA reward rises more rapidly—likely because the model inherits \textit{"parameter preferences"} from the SFT phase. As a result, it first focuses on the more easily optimized ROA, then gradually shifts attention to the PQ reward, which requires longer exploration due to the broader search space, where different edit operations may yield similar visual outcomes.
Moreover, unlike Deepseek-R1~\cite{guo2025deepseek}, JarvisArt does not display a clear “aha moment”. This absence may stem from the lack of intermediate visual feedback during the artistic reasoning process. For example, when the model makes a hypothetical retouching adjustment like highlight+5, it cannot obtain the corresponding visual result, preventing the model from validating this step's correctness within the decision-making chain. Unlike mathematical problem-solving, where each step can be validated immediately, our artistic reasoning involves numerous retouching parameters. If we perform step-wise validation for each parameter, it would require high concurrency in calling Lightroom. This is impractical due to the high computational cost and the slow training speed. Investigating step-wise visual rewards within proxy validation environments may offer a promising approach to eliciting the “aha moment”. We intend to explore in future work.

\section{Ablation Study}
\textbf{Training strategy.} 
We assess the impact of different post-training strategies by comparing model performance under three settings: 1) SFT on 50K CoT-enhanced samples, 2) GRPO-R training on 5K standard samples from scratch; and 3) GRPO-R fine-tuning basd on SFT-initial model. Rows 2–4 in Table~\ref{ablation} show that SFT yields better results than GRPO-R trained from scratch. This is likely because, without SFT to instill the basic reasoning and tool-use abilities, the GRPO-R training process must explore a significantly larger search space, thereby hindering optimization. Our combined SFT+GRPO-R strategy achieves the best results, suggesting that GRPO-R can effectively enhance the SFT-initialized model’s reasoning, tool proficiency, and generalization by expanding its exploration capacity.

\textbf{Reward design.} 
As shown in Rows 6–8 of Table~\ref{ablation}, individual reward combinations (Format+ROA or Format+PQ) result in suboptimal performance, with Format+PQ performing slightly better—possibly because PQ aligns more closely with the ultimate objective of enhancing visual quality and offers a broader optimization space to escape local optima. The full combination (Format+PQ+ROA) achieves the highest performance. This result aligns with our intuition that parameter-oriented (ROA) and perception-driven (PQ) rewards are complementary: ROA ensures parameter accuracy, while PQ maintains visual fidelity. The multi-dimensional reward system provides a balanced optimization signal, guiding the model to predict accurate edit operations while preserving high visual quality.

\begin{table*}[t]
    \centering
    \caption{Ablation studies on different training strategies and reward design.}
    \label{ablation} 
    \scalebox{0.8}{
        \setlength\tabcolsep{20pt}
        \renewcommand\arraystretch{1}
        \begin{tabular}{lccccc}
            \toprule
            \textbf{Configurations} & $\text{L1}_{\times 10^{2}}$ $\downarrow$ & $\text{L2}_{\times 10^{3}}$ $\downarrow$ & SC $\uparrow$ & PQ $\uparrow$ & O $\uparrow$ \\
            \midrule
            \multicolumn{6}{l}{\textcolor{gray!60}{Training strategy}} \\ 
            only SFT & 14.42 & 44.38 & 7.32 & 8.67 & 7.94 \\
            only RL & 17.55 & 58.19 & 6.88 & 8.13 & 7.38 \\
            SFT + RL (Ours) & \textbf{12.44} & \textbf{30.56} & \textbf{7.53} & \textbf{9.82} & \textbf{8.52} \\
            \midrule
            \multicolumn{6}{l}{\textcolor{gray!60}{Reward design}} \\ 
            Format + ROA & 14.09 & 40.36 & 7.45 & 8.77 & 8.04 \\
            Format + PQ & 13.78 & 35.41 & 7.48 & 8.92 & 8.15 \\
            Format + ROA + PQ (Ours) & \textbf{12.44} & \textbf{30.56} & \textbf{7.53} & \textbf{9.82} & \textbf{8.52} \\
            \bottomrule
        \end{tabular}
    }
\end{table*}



\section{Conclusion}

This report introduces JarvisArt, an interactive and interpretable MLLM-guided agent that integrates with 200+ Lightroom editing tools, enabling non-destructive editing on images of any-resolution. To develop this artist agent, we propose a new data generation pipeline that curates the MMArt-55K dataset, comprising 5K standard and 50K CoT-enhanced samples. Based on this dataset, we train JarvisArt using a two-stage post-training regimen: 1) CoT SFT to instill basic reasoning and tool-use abilities, and 2) GRPO-R to improve the agent’s reasoning, tool proficiency, and generalization through customized tool-use rewards: retouching operation accuracy reward for assessing the predicted editing operations, and the perceptual quality reward to evaluate the visual fidelity of the edited outputs. Furthermore, to enable seamless, automated Lightroom editing, we introduce the Agent-to-Lightroom protocol. Evaluation results from our MMArt-Bench demonstrate that our proposed algorithm significantly outperforms existing advanced image editing algorithms.

\newpage
\section*{Appendices}
Our Appendices includes the following sections:

\begin{itemize}

    \item Sec.\ref{app:MMart_dataset} Details of the MMArt Dataset.
    \begin{itemize}
        \item Statistics of the MMArt Dataset.
        \item Comparison of Existing Datasets.
        \item Data Samples of MMArt.
        \item Prompt Templates.
    \end{itemize}
    
    \item Sec.\ref{app:method_detials} Additional Method Details.
    \begin{itemize}
        \item Group Relative Policy Optimization.
        \item Details of Reward Calculation.
    \end{itemize}

    \item Sec.\ref{app:implementation} Additional Experimental Details.
    \begin{itemize}
        \item Calculation of Local Metrics.
        \item Prompts for MLLM-based Metrics.
    \end{itemize}

    \item Sec.\ref{app:more_results} Additional Experimental Results.
    \begin{itemize}
        \item Additional Quantitative Evaluation by Qwen-2.5-VL-72B.
        \item Examples of Intricate Retouching Tasks with JarvisArt.
        \item More Visual Comparisons.
        \item Comparison on MIT-FiveK.
    \end{itemize}
    
    \item Sec.\ref{app:tools_detials} Details of Retouching Tools in Lightroom.



    

\end{itemize}
\appendix






    
    

\section{Details of the MMArt dataset.}\label{app:MMart_dataset}
\subsection{Statistics of the MMArt dataset}\label{app:MMart_Statistics}
\begin{figure*}[!t]
    \centering
    \includegraphics[width=\linewidth]{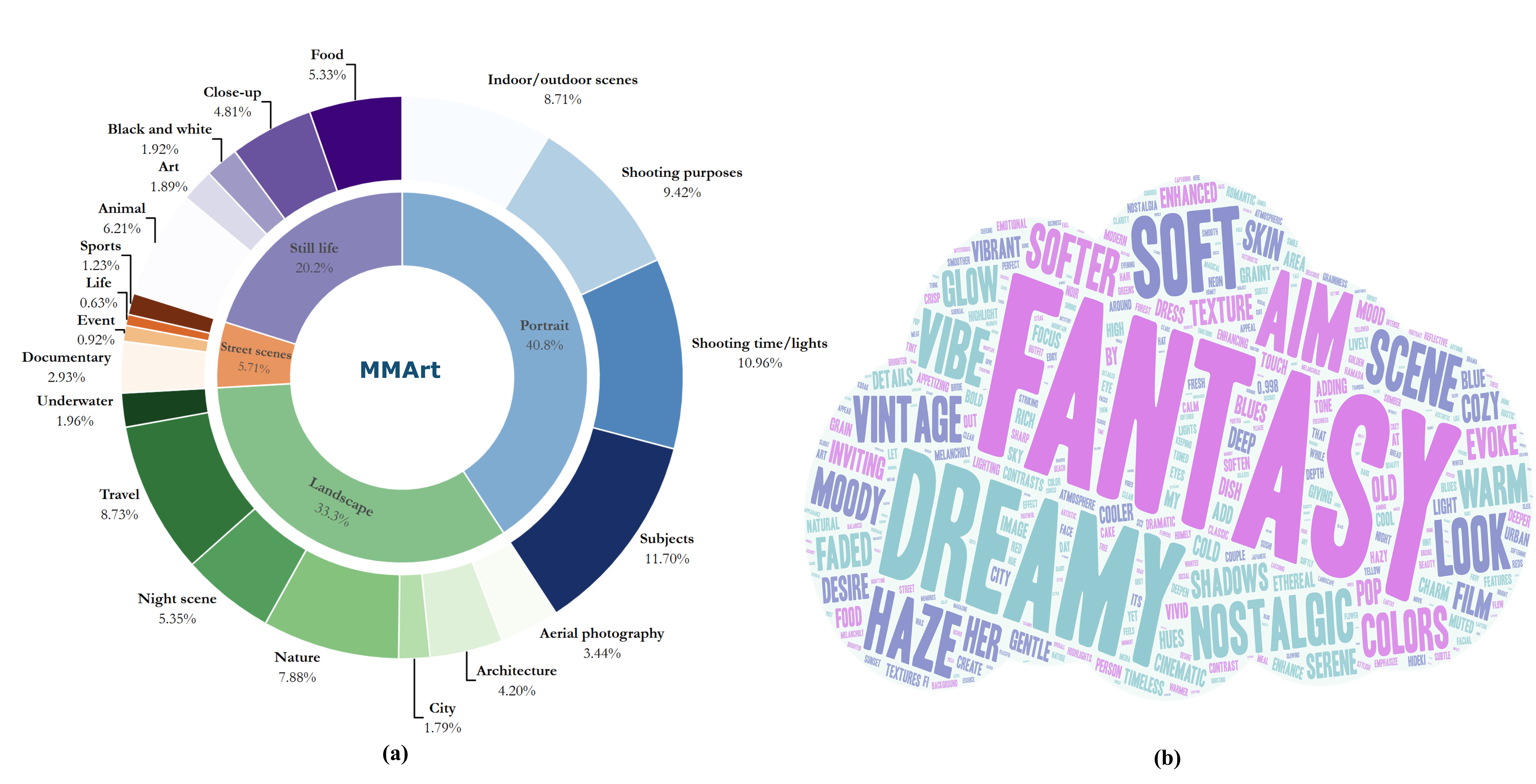}
    \caption{Statistics of the MMArt dataset. (a) The dataset is divided into four primary scenarios: portrait, landscape, street scenes, and still life, each containing a variety of subcategories. (b) A word cloud illustrates the rich linguistic diversity found in user instructions.}
    \label{fig:MMart_distribution_cloud}
    \vspace{-0.5cm}
\end{figure*}

Figure~\ref{fig:MMart_distribution_cloud}(a) illustrates the composition and distribution of scenarios in our MMArt dataset. The dataset is structured into four major scene categories that reflect common real-world photo retouching contexts: portrait (40.8\%, including shooting purposes, shooting time/lights, subjects, and indoor/outdoor scenes), landscape (33.3\%, comprising nature, city, aerial photography,travel, underwater, night scene and architecture), street scenes (5.71\%, including sports, life, event and documentary), and still life (20.2\%, encompassing food, close-up scenes, black/white photography, art and animals). Each major category contains a diverse set of subcategories, ensuring comprehensive coverage and representativeness. Furthermore, Figure~\ref{fig:MMart_distribution_cloud}(b) displays a word cloud of user instructions, highlighting the linguistic diversity of the instructions.

\subsection{Comparison of Existing Datasets}\label{app:dataset_properties}
Table~\ref{tab:MMArt_properties_comparison} presents a comparison between our MMArt dataset and existing image editing datasets. MMArt is designed with the following key properties to facilitate advanced research in image retouching:
\begin{itemize}
    \item \textbf{Real Images:} All samples in MMArt are real photographs, ensuring the dataset's authenticity and practical value for real-world applications.
    \item \textbf{Diverse User Instructions:} Each image is paired with detailed user instructions, capturing a wide variety of editing intentions and reflecting the diversity of natural language expressions.
    \item \textbf{Flexible Resolution:} MMArt supports images of any resolution, including high-resolution samples, making it suitable for both research and practical deployment scenarios.
    \item \textbf{Chain-of-Thought (CoT) Annotations:} The dataset provides CoT reasoning annotations, which help to reveal the underlying logic and step-by-step process of user intent understanding and image editing.
    \item \textbf{Lightroom Retouching Configuration:} For every sample, MMArt includes comprehensive Lightroom parameter configurations, allowing for non-destructive, reproducible, and transparent image editing.
\end{itemize}
These properties make MMArt a high-quality, flexible, and richly annotated resource for the development and evaluation of advanced image retouching techniques.



\begin{table}[t!]
\centering
\caption{Comparison of MMArt and existing retouching datasets in terms of data properties.}
\label{tab:MMArt_properties_comparison}
\scalebox{0.68}{
\begin{tabular}{lccccccc}
\toprule
\textbf{Property} & \textbf{InstructP2P~\cite{brooks2023instructpix2pix}} & \textbf{MagicBrush~\cite{zhang2023magicbrush}} & \textbf{UltraEdit~\cite{zhao2024ultraedit}} & \textbf{MGIE~\cite{fu2024mgie}} & \textbf{HQEdit~\cite{hui2024hq}} & \textbf{FiveK~\cite{fivek}} & \textbf{MMArt} \\
\midrule
Real Image? & \xmark & \cmark & \cmark & \cmark & \xmark & \cmark & \cmark \\
User Instructions?& \cmark & \cmark & \cmark & \cmark & \cmark & \xmark & \cmark \\
Any Resolution? & \xmark & \xmark & \xmark & \xmark & \xmark & \cmark & \cmark \\
High Resolution? & \xmark & \xmark & \xmark & \xmark & \cmark & \cmark & \cmark \\
CoT Annotations? & \xmark & \xmark & \xmark & \xmark & \xmark & \xmark & \cmark\\ 
Lightroom Configuration? & \xmark & \xmark & \xmark & \xmark & \xmark & \cmark & \cmark \\
\bottomrule
\end{tabular}
}
\end{table}

\begin{figure*}[!t]
    \centering
    \includegraphics[width=\linewidth]{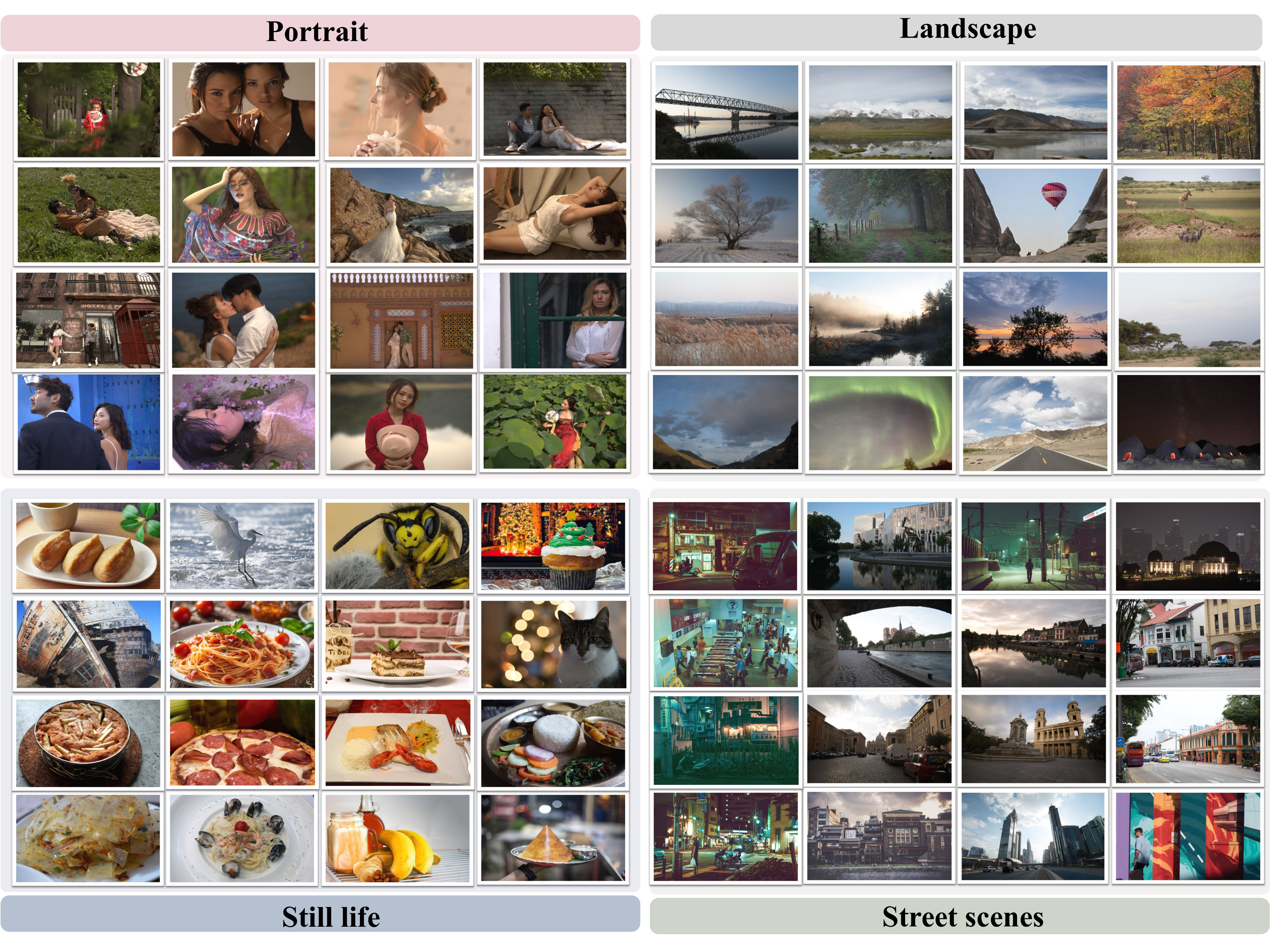}
    \caption{Visual examples to demonstrate the diversity of the proposed dataset.}
    \label{fig:MMart_image_smapes}
\end{figure*}

\subsection{Data Samples of MMArt}\label{app:data_example}
\begin{figure*}[htbp]
    \centering
    \includegraphics[width=\linewidth]{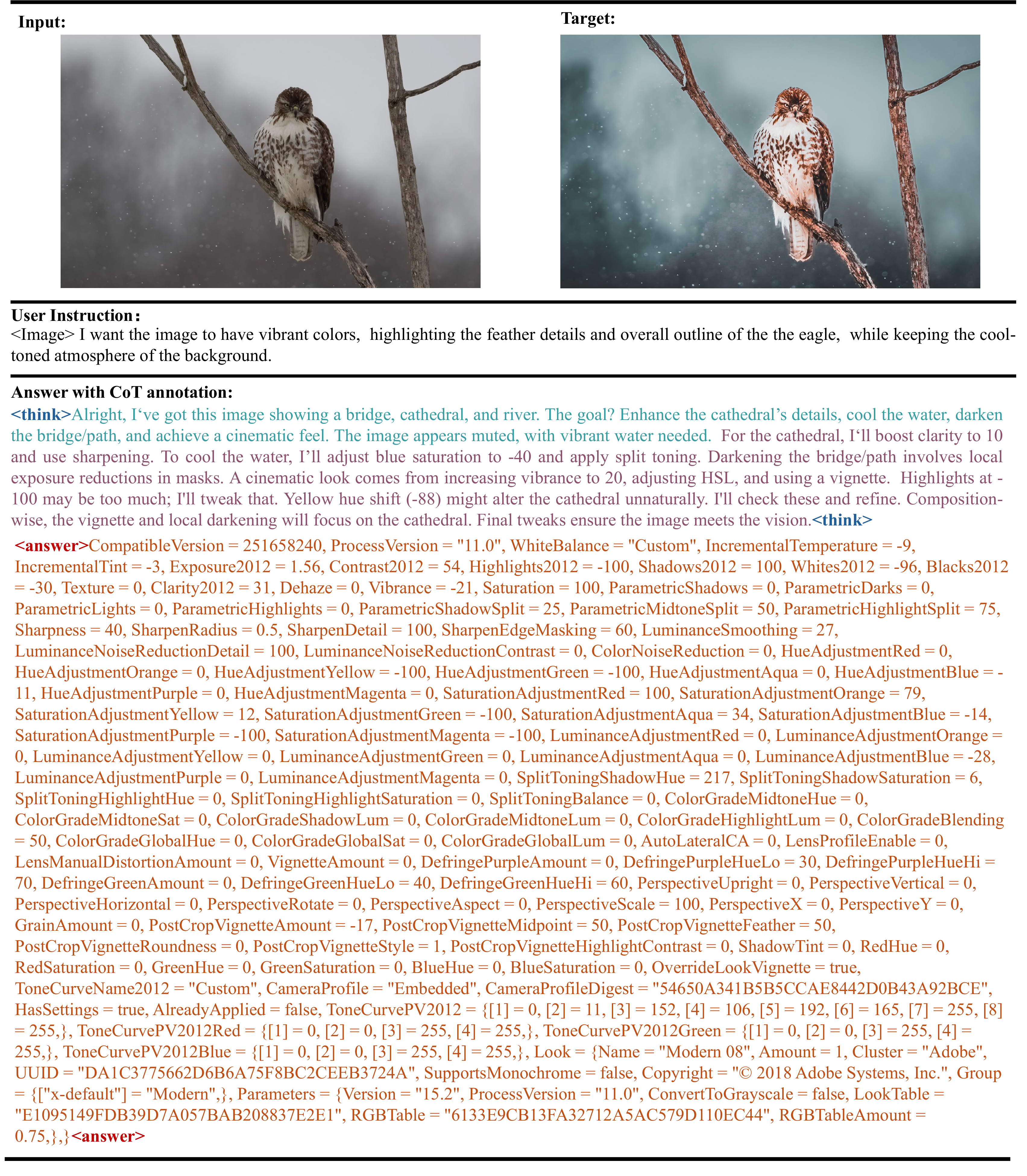}
    \caption{Examples of MMArt data annotated with Chain-of-Thought (CoT) reasoning.}
    \label{fig:Insrtuction_COT_data_sample}
\end{figure*}

\begin{figure*}[htbp]
    \centering
    \includegraphics[width=\linewidth]{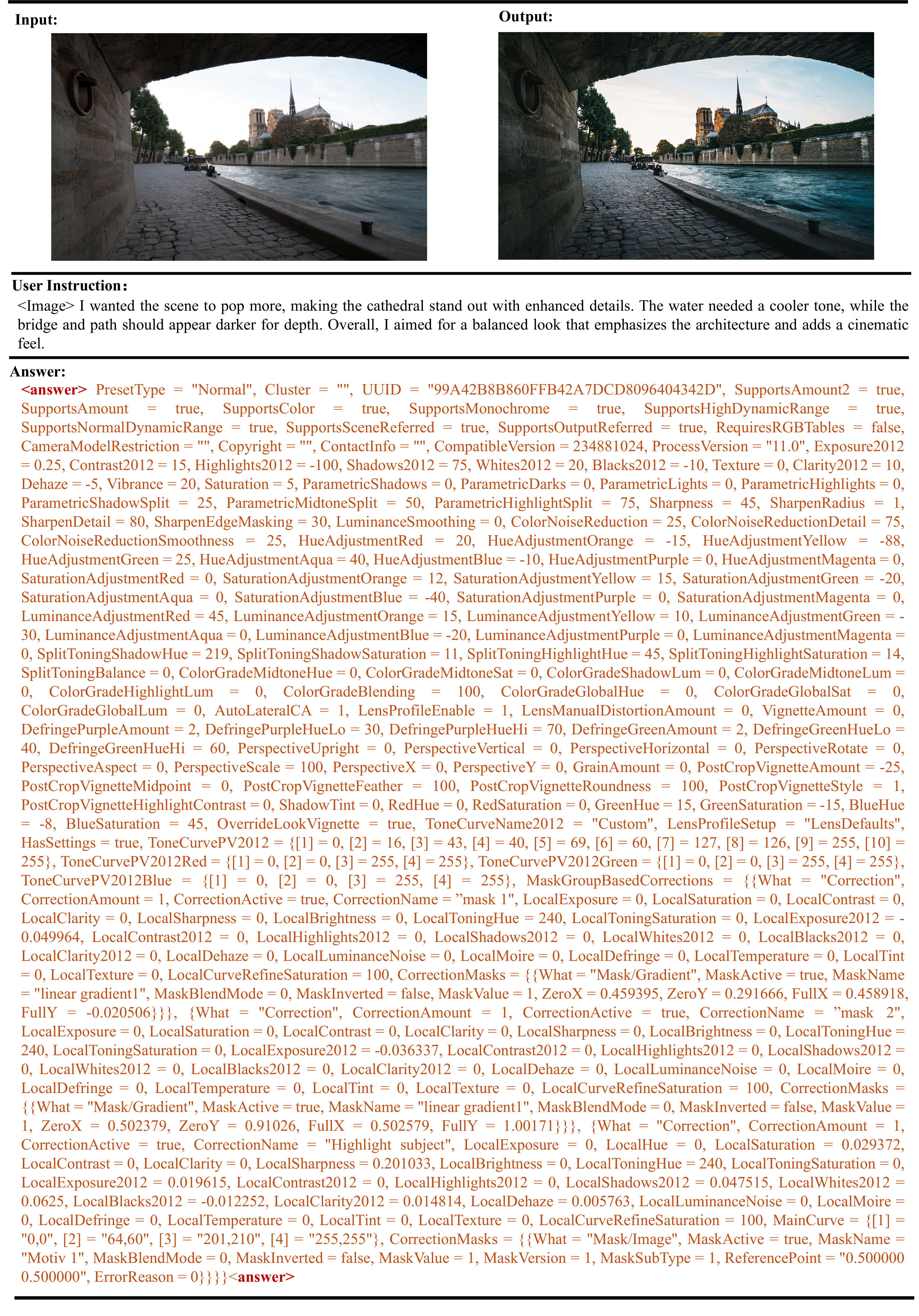}
    \caption{Data samples from MMArt with standard instructions.}
    \label{fig:Insrtuction_data_sample}
\end{figure*}

The diversity of collected photos is shown in Figure~\ref{fig:MMart_image_smapes}. Moreover,  Figure~\ref{fig:Insrtuction_COT_data_sample} demonstrates MMArt samples with Chain-of-Thought (CoT) reasoning, while Figure ~\ref{fig:Insrtuction_data_sample} shows standard examples without CoT annotations. 

\subsection{Prompt Templates}\label{app:recommed_temple}
\begin{figure*}[htbp]
    \centering
    \includegraphics[width=\linewidth]{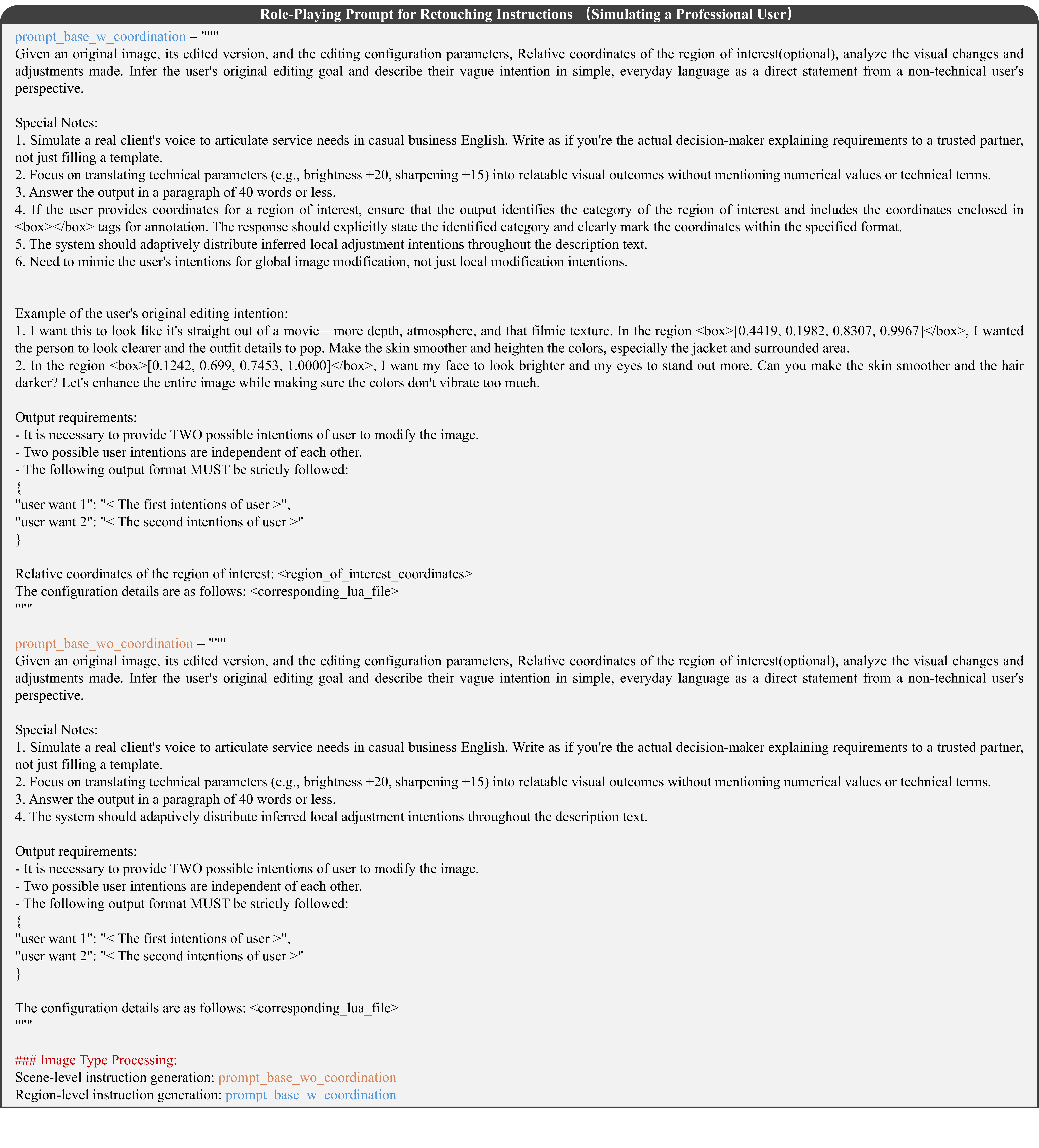}
    \caption{Prompt for simulating the professional user instructions
    }
    \label{prompt_user_professional}
\end{figure*}

\begin{figure*}[htbp]
    \vspace{-2cm}
    \centering
    \includegraphics[width=\linewidth]{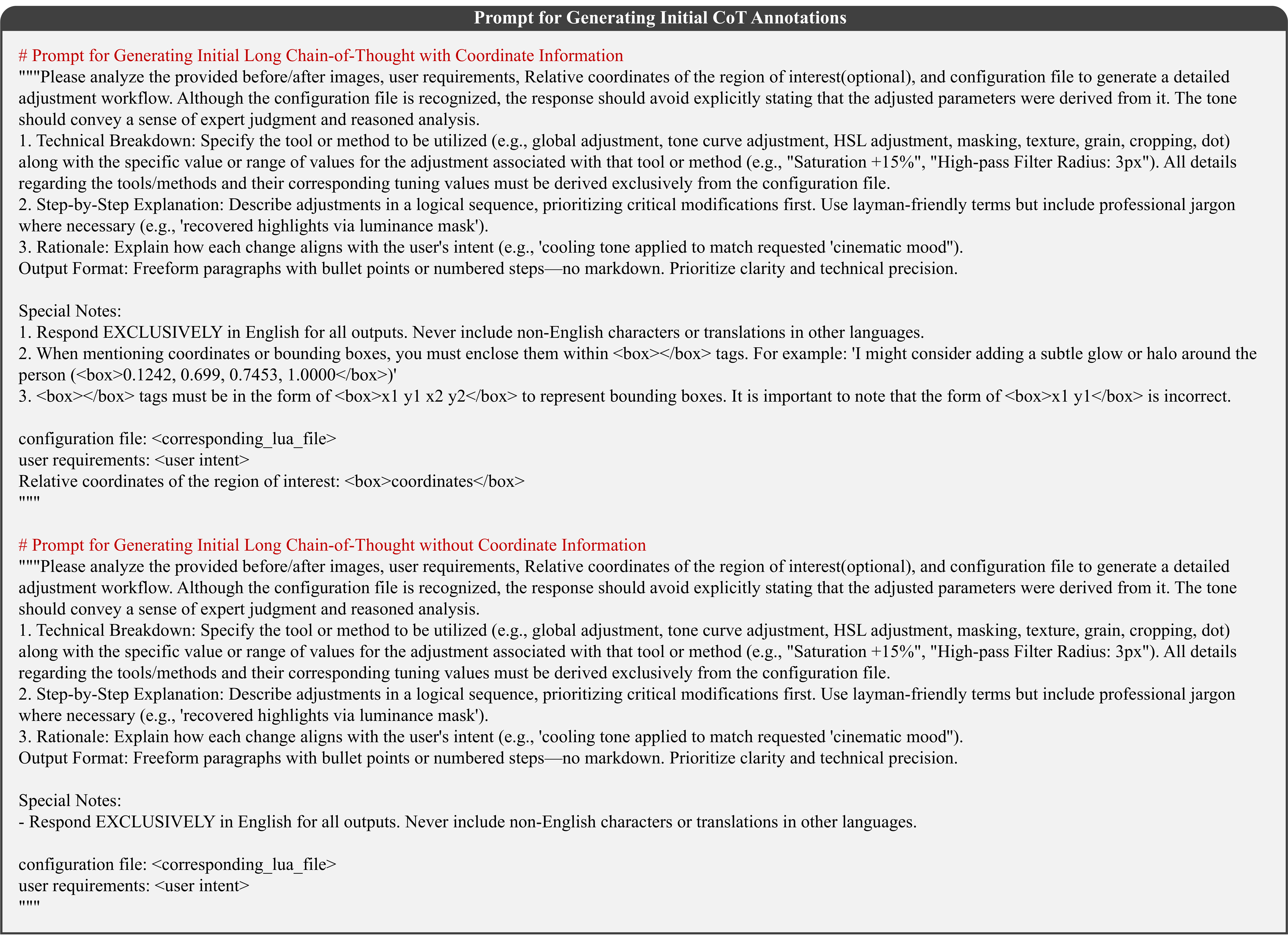}
    \caption{Prompt for generating the initial Chain-of-Thought (COT) annotations.
    }
    \label{long_cot_prompt}
\end{figure*}

\begin{figure*}[htbp]
    \centering
    \includegraphics[width=\linewidth]{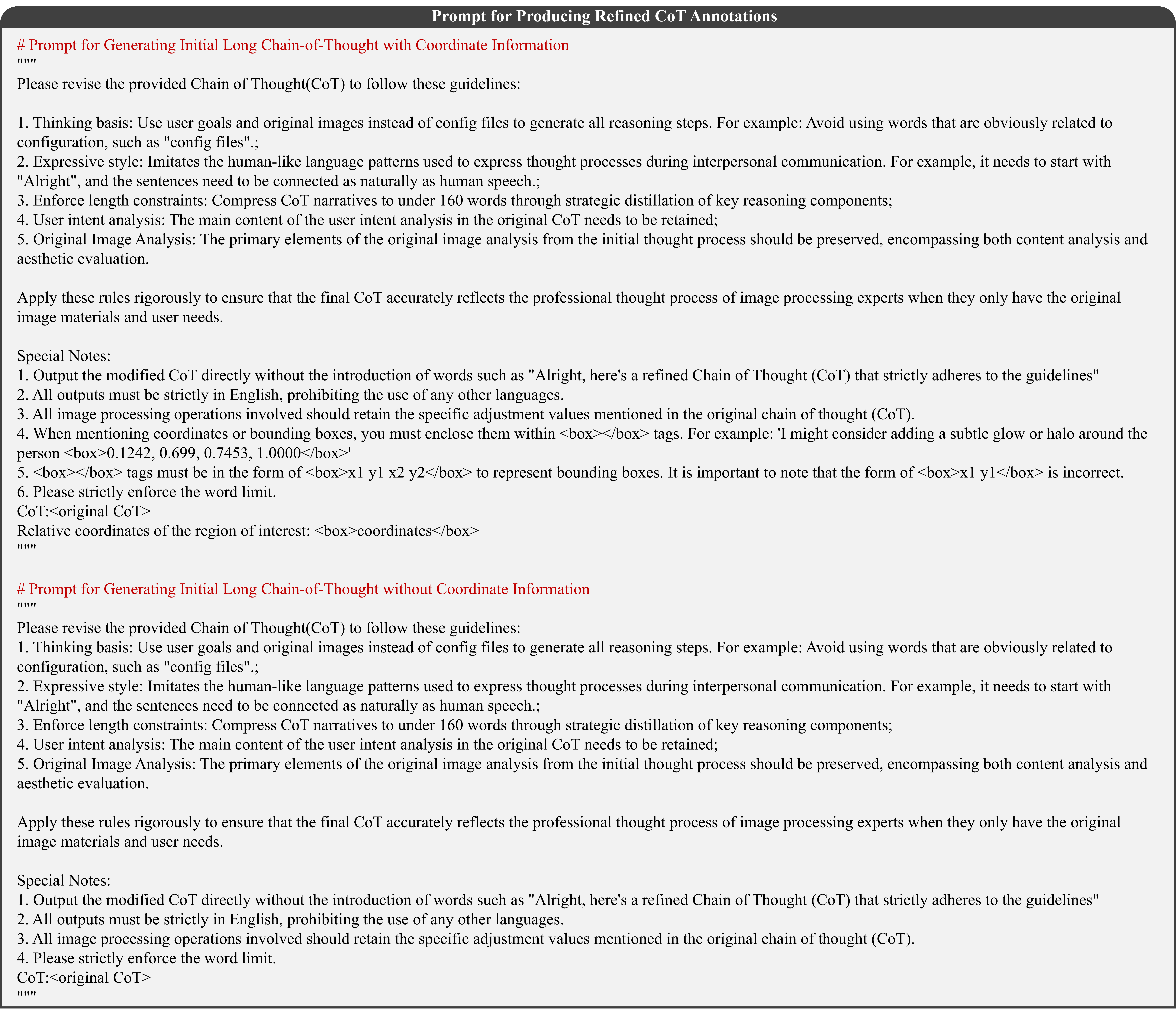}
    \caption{Prompt for generating the refined Chain-of-Thought (COT) annotations.
    }
    \label{short_cot_prompt}
\end{figure*}

The prompt templates utilized throughout the various stages of MMArt are summarized here—Aesthetic Preset Recommendation (Figure~\ref{prompt_preset_recom}), User Instruction Simulation (Figures~\ref{prompt_user_professional} and \ref{prompt_user_casual}), and Chain-of-Thought Data Construction (Figure~\ref{long_cot_prompt} and \ref{short_cot_prompt}).

\section{Additional Method Details}\label{app:method_detials}
\subsection{Group Relative Policy Optimization}\label{app:GRPO_detials}
 In GRPO, given a task question, the model generates a set of \( N \) potential responses \(\{O_1, O_2, \ldots, O_N\}\). Each response is evaluated by taking the corresponding actions and computing its reward \(\{R_1, R_2, \ldots, R_N\}\). Unlike PPO, which relies on a single reward signal and a critic to estimate the value function, GRPO normalizes these rewards to calculate the relative advantage of each response. The relative quality \( A_i \) of the i-th response is computed as
\begin{equation*}
A_i = \frac{r_i - \text{Mean}(\{r_1, r_2, \ldots, r_N\})}{\text{Std}(\{r_1, r_2, \ldots, r_N\})},
\end{equation*}
where \text{Mean} and \text{Std} represent the mean and standard deviation of the rewards, respectively. This normalization step ensures that responses are compared within the context of the group, allowing GRPO to better capture nuanced differences between candidates. Policy updates are further constrained by minimizing the KL divergence between the updated and reference models, ensuring stable RL learning. Refer to \cite{guo2025deepseek,shao2024deepseekmath} for more details.

\subsection{Details of Reward Calculation}\label{app:rewards_calculation}
The parameter value matching function $S_k(\cdot)$ for each parameter $k$ is determined based on its specific type. Let $V_{k}^{pre}$ and $V_{k}^{tgt}$ denote the predicted and ground truth values for the $k$-th parameter, respectively. For notational simplicity, we omit the subscript $k$ in the following formulas. The calculation proceeds as follows:
\begin{itemize}
    \item \textbf{Scalar Parameters.}  
    For scalar parameters such as exposure or contrast, the matching function $S$ is defined as:
    \begin{equation*}
        S = \max\left(0,1 - \frac{|V^{pre} - V^{tgt}|}{V_{max} - V_{min}} \right) \in [0,1],
    \end{equation*}
    where $|\cdot|$ represents the absolute error between the predicted and ground truth values.

    \item \textbf{Linear Gradient Masks.}  
    We assess the similarity between predicted and target linear gradient masks by measuring the distances between their start points $p_s = (x_s,y_s)$ and end points $p_e = (x_e,y_e)$, with coordinates normalized to [0,1] for resolution invariance. The similarity score is computed as:
    \begin{equation*}
        S = \max\left(0, 1 - \|p_s^{\text{pre}} - p_s^{\text{tgt}}\| - \|p_e^{\text{pre}} - p_e^{\text{tgt}}\|\right) \in [0,1],
    \end{equation*}
    where $\|\cdot\|$ denotes Euclidean distance.

    \item \textbf{Radial Gradient Masks.}  
    We measure similarity between predicted and target radial gradient masks using three geometric parameters: center position $c = (x, y)$, scale factors $(W, H)$, and rotation angle $\theta$. Center point similarity is given by:
    \begin{equation*}
        S_{center} = \max(0, 1 - 2\cdot \|c^{pre} - c^{tgt}\|) \in [0,1],
    \end{equation*}
    where $c^{pre}$ and $c^{tgt}$ are normalized to [0,1]. Further, scaling similarity compares width/height ratios:    
    \begin{equation*}
        S_{scale} = \max(0, 1 - |W^{pre}/W^{tgt} - 1| - |H^{pre}/H^{tgt} - 1|) \in [0,1],
    \end{equation*}
    The angle numerical value similarity is defined by:
    \begin{equation*}
        S_{angle} = \max\left(0, 1 - \frac{|\theta^{pre} - \theta^{tgt}|}{\theta_{max} - \theta_{min}} \right) \in [0,1],
    \end{equation*}
    The final similarity score combines these components as follows: 
    \begin{equation*}
        S = 0.4 \cdot S_{center} + 0.4 \cdot S_{scale} + 0.2 \cdot S_{angle} \in [0,1].
    \end{equation*}

    \item \textbf{Object Masks.}  
    For object masks, the similarity score $S$ is defined as the Intersection-over-Union (IoU) between the predicted $B^{pre}$ and ground truth $B^{tgt}$ bounding boxes. Each box is parameterized as $[x_1, y_1, x_2, y_2]$. The similarity score is computed as:
    \begin{equation*}
        \text{S} = \text{IoU}(B^{pre},B^{tgt}) \in [0,1],
    \end{equation*}
    where higher values indicate better alignment, with $S = 1$ denoting perfect overlap and $S = 0$ indicating no intersection.

    \item \textbf{Portrait Masks.}  
    In portrait masks, the model predicts different special category IDs to denote distinct regions, such as ID=0 for face, ID=1 for hair, ID=2 for eyes, ID=3 for skin, etc. The matching score $S$ is defined as follows:
    \begin{equation*}
        S = 
        \begin{cases}
            1, & \text{if the predicted and target category IDs coincide,} \\
            0, & \text{otherwise}. \\
        \end{cases}
    \end{equation*}

    \item \textbf{Color Range Masks.}  
    To evaluate color range mask similarity, we sample $N$ representative points from both predicted and target color distributions and compute the mean CIEDE2000 color difference $\Delta E_{100}$ in LAB color space. The similarity score is given by:
    \begin{equation*}
        S = \max\left(0, 1 - \frac{1}{N}\sum_{n=1}^N \Delta E_{100}(c_n^{pre}, c_n^{tgt})\right) \in [0,1],
    \end{equation*}
    where $c_n^{pre}$ and $c_n^{tgt}$ denote the $n$-th sampled colors from the predicted and ground-truth distributions, respectively.

    \item \textbf{Luminance Range Masks.}  
    To evaluate luminance range mask similarity, we compare the predicted and target luminance extremes by computing their absolute differences. The similarity score is defined as:
    \begin{equation*}
        S = \max\left(0, 1 - \frac{|l_{\min}^{pre}-l_{\min}^{tgt}| + |l_{\max}^{pre}-l_{\max}^{tgt}|}{2(l_{\max}^{tgt}-l_{\min}^{tgt})} \right) \in [0,1],
    \end{equation*}
    where the denominator normalizes by the target luminance range to ensure scale invariance.
\end{itemize}

\section{Additional Experimental Details}\label{app:implementation}
\subsection{Calculation of Local Metrics}\label{app:metric_calculate}


To evaluate the model's effectiveness in localized regions, we compute six metrics—$\mathrm{L1}^{RC}$, $\mathrm{L2}^{RC}$, $\text{SC}^{RC}$, $\text{PQ}^{RC}$, and $\text{O}^{RC}$—using human-centric masks from the portrait subset of MMArt-Bench. For $\mathrm{L1}^{RC}$ and $\mathrm{L2}^{RC}$, inspired by PPR10K~\cite{liang2021ppr10k}, given an image $I$ with resolution $H \times W$, we define a weighting matrix $W_I = [w_{i,j}] \in \mathbb{R}^{H \times W}$, where $w_{i,j} = 1$ for human regions and $w_{i,j} = \alpha (\alpha \leq 1)$ for background regions, with $\alpha$ empirically set to 0.5. For instance, the human-centric L1 difference metric is expressed as:
\[
\ \mathrm{L1}^{RC} = | W_I \circ I^{pred} - W_I \circ I^{tgt} |,
\]
where $I^{\text{pred}}$ and $I^{\text{tgt}}$ are the predicted and target images, respectively, and $\circ$ denotes element-wise multiplication. The $\mathrm{L2}^{RC}$ metric is defined in a similar manner. For $\text{SC}^{RC}$, $\text{PQ}^{RC}$, and $\text{O}^{RC}$, with $\alpha$ empirically set to 0, we focus solely on the mask region of the edited image and prompt the MLLM to emphasize local adjustments.

\subsection{Prompt for MLLM-based Metrics}\label{app:mllm_metrics}
As shown in Figure~\ref{prompt_metrics}, we present the evaluation prompts utilized for both scene-level and region-level assessments of the Semantic Consistency (SC) and Perceptual Quality (PQ) metrics. Notably, the overall score is calculated as $\text{O} = \sqrt{\text{SC} \times \text{PQ}}$.

\section{Additional Experimental Results}\label{app:more_results}

\subsection{Additional Quantitative Evaluation by Qwen2.5-VL-72B}
As shon in Table~\ref{app:table}, to further evaluate MLLM-based metrics, we conducted an additional quantitative analysis using Qwen2.5-VL-72B~\cite{bai2025qwen2}. Our findings suggest that such metrics may be unreliable, struggle to effectively reflect a model's instruction-following capability. Despite this, our model demonstrates instruction-following performance comparable to that of contemporary SOTA closed-source model GPT-4o, while achieving a significant improvement in content fidelity.

\begin{table*}[t!]
    \centering
    \setlength{\abovecaptionskip}{0.1cm} 
    \setlength{\belowcaptionskip}{-0.6cm}
    \caption{Quantitative evaluation on MMArt-Bench. We highlight the \colorbox{lightpeach!90}{best} and \colorbox{teasergree!20}{second-best} instruction-based results. SC, PQ, and O refer to the metrics evaluated by Qwen2.5-VL-72B~\cite{bai2025qwen2}. The $RC$ means the metric calculated on specific mask region. }\label{app:table}
    \scalebox{0.8}{
    \setlength\tabcolsep{2pt}
    \renewcommand\arraystretch{1}
    \begin{tabular}{lcccccccccccc}
    \toprule
    ~ & ~ & \multicolumn{5}{c}{Scene-level} & \multicolumn{5}{c}{Region-level} \\ 
    \cmidrule(lr){3-7} \cmidrule(lr){8-12} 
    \multirow{-2}{*}{Method} & \multirow{-2}{*}{Instruction} & $\text{L1}_{\times 10^{2}}$ $\downarrow$ & $\text{L2}_{\times 10^{3}}$ $\downarrow$ & SC $\uparrow$ & PQ $\uparrow$  & O $\uparrow$ & $\mathrm{L}1_{\times 10^{2}}^{RC}$ $\downarrow$ & $\mathrm{L}2_{\times 10^{3}}^{RC}$ $\downarrow$  & $\text{SC}^{RC}$ $\uparrow$ & $\text{PQ}^{RC}$ $\uparrow$ & $\text{O}^{RC}$ $\uparrow$ \\
    \midrule
    RSFNet~\cite{ouyang2023rsfnet}    & \xmark  & 11.62  & 26.38 & - & - & - & 8.80 & 13.70 & - & - & - &  \\
    3DLUT~\cite{zeng2020learning}     & \xmark & 11.51  & 26.00 & - & - & - & 8.34 & 12.26 & - & - & - &  \\
    \midrule
    \midrule
    InstructPix2Pix~\cite{brooks2023instructpix2pix}  & \cmark  & \second{15.62}  & \second{47.26} & 6.17 & 5.81 & 5.47 & 12.48 & \second{32.69} & 4.67 & 3.85 & 3.64 \\
    MagicBrush~\cite{zhang2023magicbrush} & \cmark & 18.31  & 64.76 & 3.93 & 2.25 & 2.44 & \second{12.44} & 32.93 & 2.80 & 2.15 & 2.01  \\
    OmniGen~\cite{xiao2024omnigen}  & \cmark  & 28.40  & 132.82 & 4.14 & 2.16 & 2.70 & 24.85 & 106.81 & 3.80 & 3.85 & 3.67  \\
    VARGPT-v1.1~\cite{zhuang2025vargpt}  & \cmark  & 27.04  & 126.26 & 1.27 & 0.17 & 0.29 & 23.59 & 105.86 & 0.09 & 0.02 & 0.03  \\
    Step1X-Edit~\cite{liu2025step1x}  & \cmark  & 24.17  & 105.14 & 7.03 & 4.94 & 5.71 & 15.27 & 44.93 & 7.50 & 6.89 & 7.11  \\
    \midrule
    Gemini-2-Flash~\cite{team2023gemini} & \cmark  & 23.06  & 90.96 & \second{7.65} & 6.77 & \second{7.00} & 16.74 & 53.76 & 7.33 & 7.17 & 7.19 \\
    GPT-4o~\cite{hurst2024gpt}  & \cmark  & 22.77  & 91.79 & \best{8.52} & \second{7.37} & \best{7.85} & 15.67 & 47.60 & \best{8.07} & \second{7.87} & \best{7.95}  \\
    \midrule

   \textbf{JarvisArt}& \cmark & \best{12.66} & \best{31.88} &  6.19 &  \best{8.51} & 6.67 & \best{7.75} & \best{12.38} & \second{7.54} & \best{8.46} &  \second{7.91} \\
    \bottomrule
    \end{tabular}}
\end{table*}

\begin{table*}[!t]
  \centering
  \setlength{\abovecaptionskip}{0.1cm} 
  \caption{Quantitative evaluation on MIT-FiveK~\cite{fivek}. We highlight the \colorbox{lightpeach!90}{best} and \colorbox{teasergree!20}{second-best} instruction-based results. SC, PQ, and O refer to the metrics evaluated by Gemini-2-Flash.}\label{tab:five5K}
  \scalebox{1}{
  \setlength\tabcolsep{8.5pt}
  \renewcommand\arraystretch{1}
  \begin{tabular}{lcccccc}
  \toprule
  Method & Instruction & $\text{L1}_{\times 10^{2}}$ $\downarrow$ & $\text{L2}_{\times 10^{3}}$ $\downarrow$ & SC $\uparrow$ & PQ $\uparrow$  & O $\uparrow$  \\
  \midrule
  InstructPix2Pix~\cite{brooks2023instructpix2pix}  & \cmark  & \second{16.23}  & \second{49.54} & 6.36 & 8.34 & 7.15  \\
  MagicBrush~\cite{zhang2023magicbrush} & \cmark & 17.29  & 53.45 & 4.92 & 5.50 & 4.95  \\
  OmniGen~\cite{xiao2024omnigen}  & \cmark  & 28.53  & 128.59 & 3.12 & 2.48 & 2.57  \\
  VARGPT-v1.1~\cite{zhuang2025vargpt}  & \cmark  & 26.96  & 117.16 & 2.94 & 2.00 & 2.29  \\
  Step1X-Edit~\cite{liu2025step1x}  & \cmark  & 22.08  & 91.72 & 7.20 & 8.48 & 7.69  \\
  \midrule
  Gemini-2-Flash~\cite{team2023gemini} & \cmark  & 18.69  & 61.27 & \second{7.86} & 9.22 & 8.47  \\
  GPT-4o~\cite{hurst2024gpt}  & \cmark  & 21.49  & 78.11 & \best{8.72} & \second{9.76} & \best{9.22}  \\
  \midrule
 \textbf{JarvisArt}& \cmark & \best{12.98} & \best{30.05} &  7.36 &  \best{9.82} & \second{8.48}  \\
  \bottomrule
  \end{tabular}}
\end{table*}
\subsection{Examples of Intricate Retouching Tasks with JarvisArt}
\begin{figure*}[!t]
    \centering
    \includegraphics[width=\linewidth]{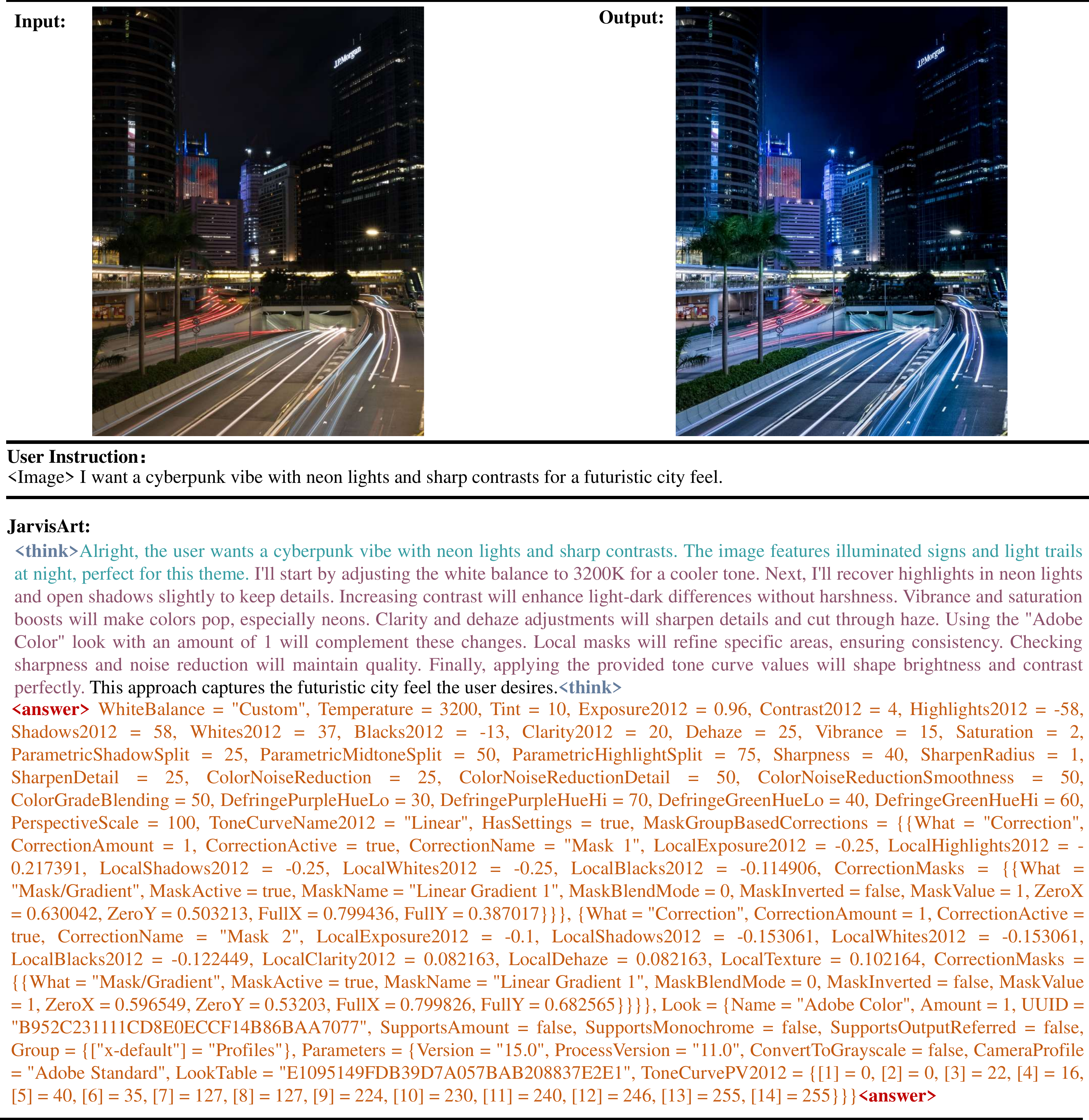}
\caption{An example of JarvisArt empowering users to achieve interactive and interpretable editing, transforming their ambiguous intentions into artistic visual outcomes.}
    \label{fig:challenge_samples1}
\end{figure*}

\begin{figure*}[!t]
    \centering
    \includegraphics[width=\linewidth]{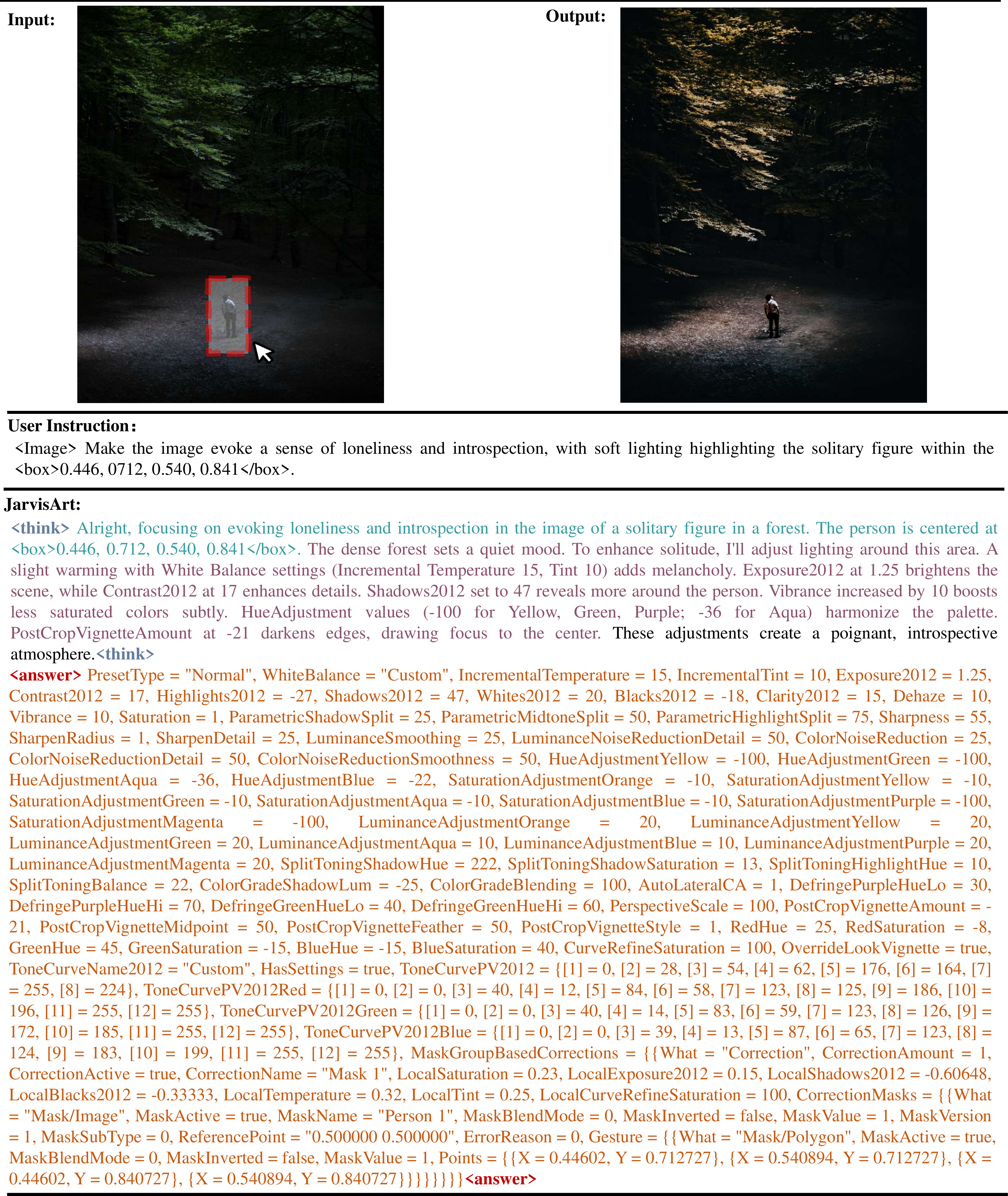}
\caption{An example of JarvisArt empowering users to achieve interactive and interpretable editing, transforming their ambiguous intentions into artistic visual outcomes.}
    \label{fig:challenging_samples2}
\end{figure*}

\begin{figure*}[!t]
  \setlength{\abovecaptionskip}{0.1cm} 
  \centering
  \includegraphics[width=1\linewidth]{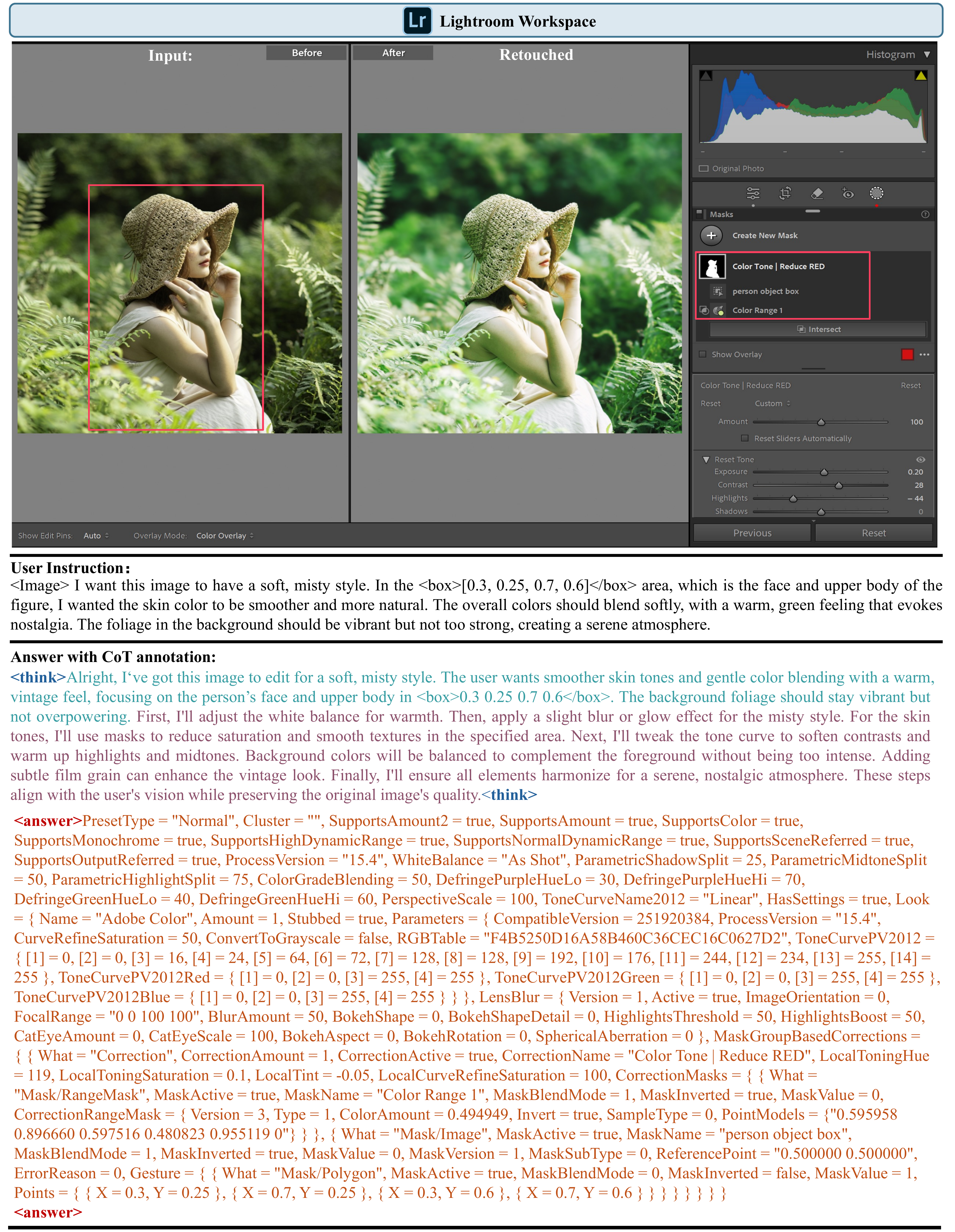}
  \caption{Editing results with JarvisArt are generated under complex prompts, with all retouching operations performed in a Lightroom environment, allowing for iterative adjustments.}
  \label{fig:challenging_samples3}
\end{figure*}

\begin{figure*}[!t]
  \setlength{\abovecaptionskip}{0.1cm} 
  \centering
  \includegraphics[width=1\linewidth]{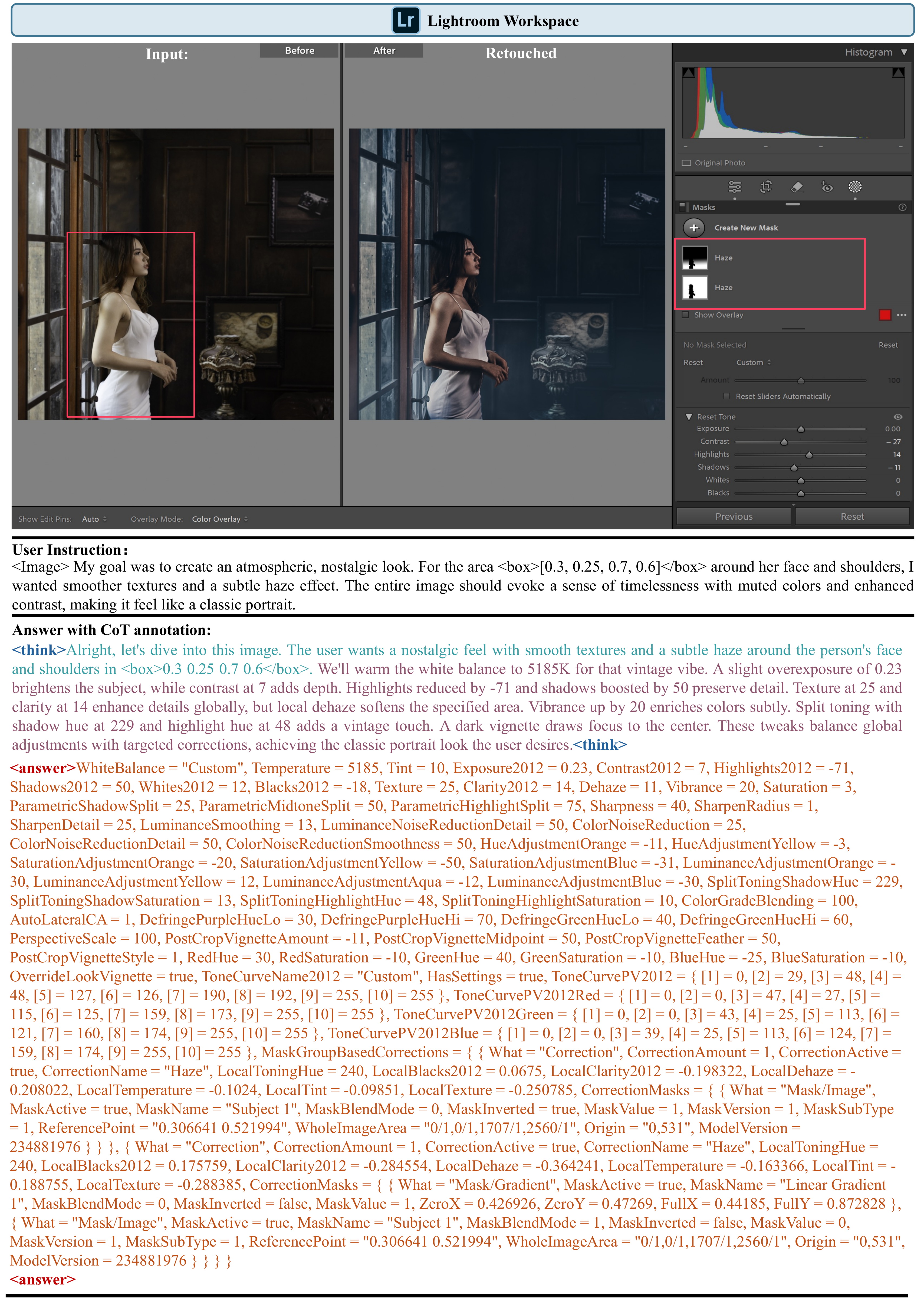}
  \caption{Editing results with JarvisArt are generated under complex prompts, with all retouching operations performed in a Lightroom environment, allowing for iterative adjustments.}
  \label{fig:challenging_samples4}
\end{figure*}

Figures~\ref{fig:challenge_samples1}-\ref{fig:challenging_samples4} present the challenging retouching examples, which involve both global and local editing demands, as well as vague user instructions. JarvisArt excels in understanding these ambiguous intentions, applying modifications at both the scene and region levels, and delivering visually effective results in the final images.

\subsection{More Visual Comparisons}\label{app:more_visual_comparation}
\begin{figure*}[!t]
    \centering
    \includegraphics[width=\linewidth]{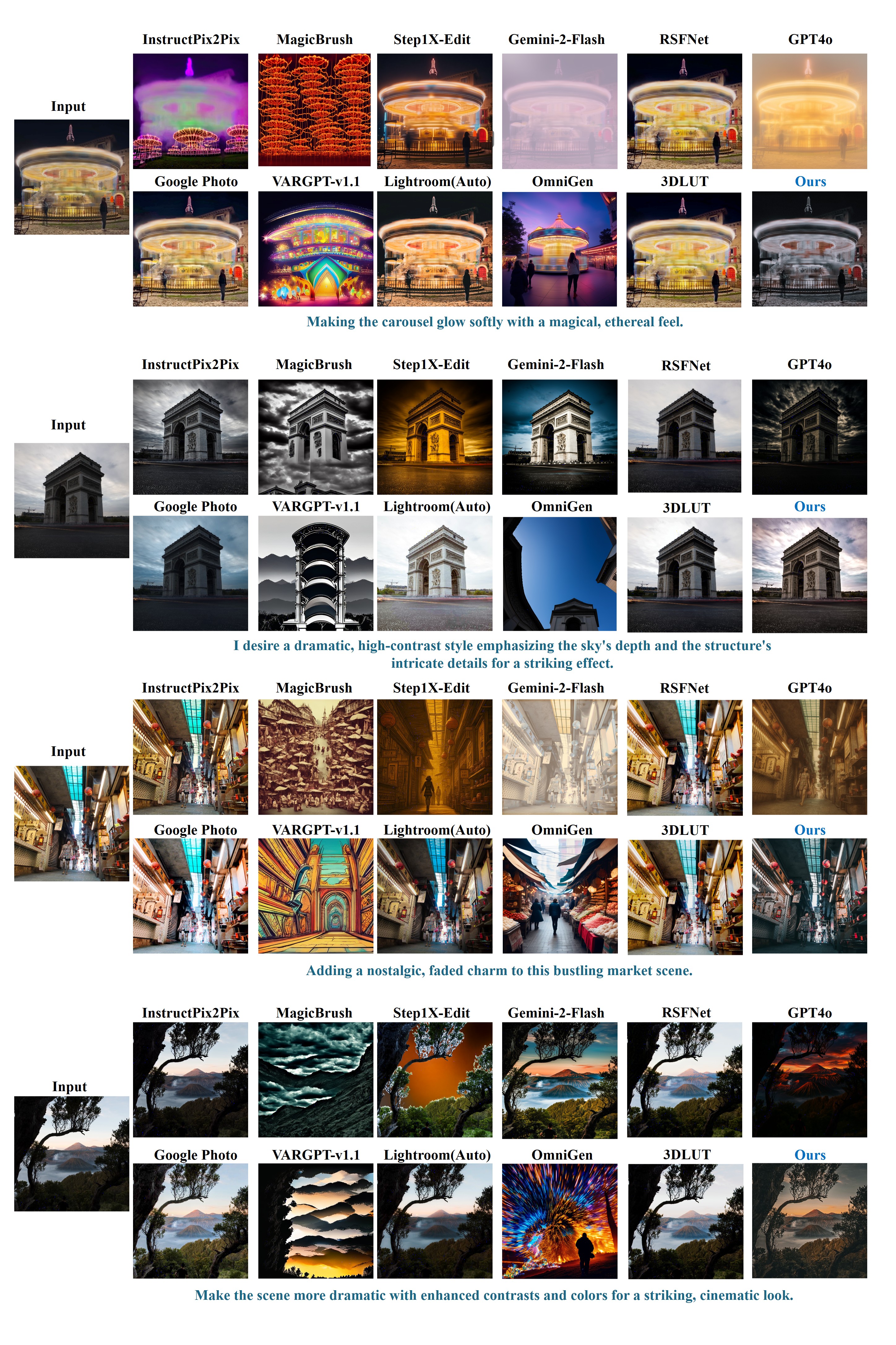}
    \caption{Visual comparisons of all state-of-the-art editing methods alongside two automatic retouching modes from commercial software.}

    \label{fig:app_viusal_reuslts1}
\end{figure*}

\begin{figure*}[htbp]
    \centering
    \includegraphics[width=\linewidth]{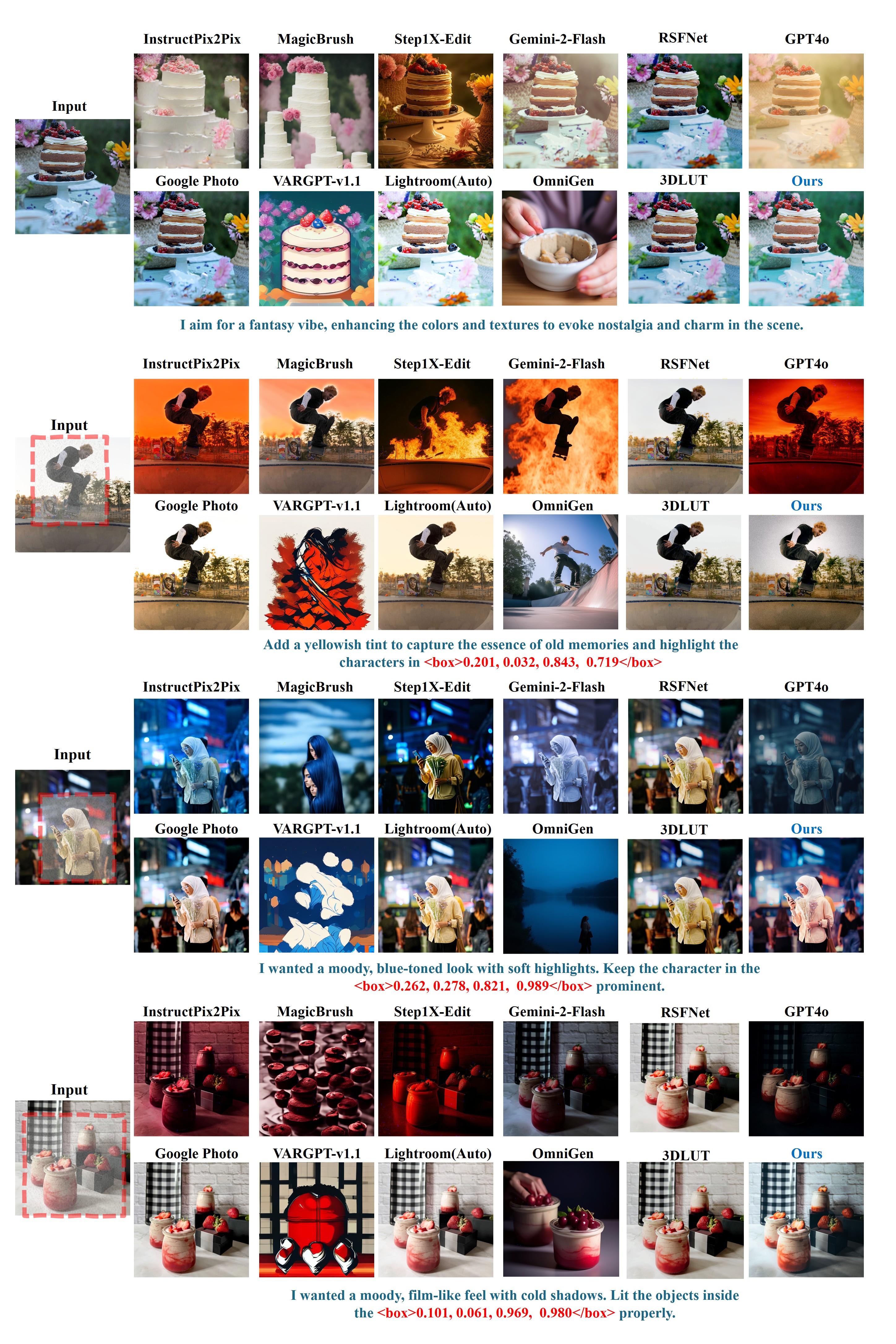}
    \caption{Visual comparisons of all state-of-the-art editing methods alongside two automatic retouching modes from commercial software.}
    \label{fig:app_viusal_reuslts2}
\end{figure*}

Figures~\ref{fig:app_viusal_reuslts1}-\ref{fig:five5k_results2} present additional photo retouching results from the MMArt-Bench, highlighting the superiority of JarvisArt in terms of instruction adherence, content fidelity, and visual appeal, while also achieving the style most similar to the target image. Notably, we also include a visual comparison with two commercial editing tools: Adobe Lightroom and Google Photo' auto-retouching modes.

\subsection{Comparison on MIT-FiveK}
\begin{figure*}[!t]
  \vspace{-1.5cm}
  \setlength{\abovecaptionskip}{-0.1cm} 
  \centering
  \includegraphics[width=0.85\linewidth]{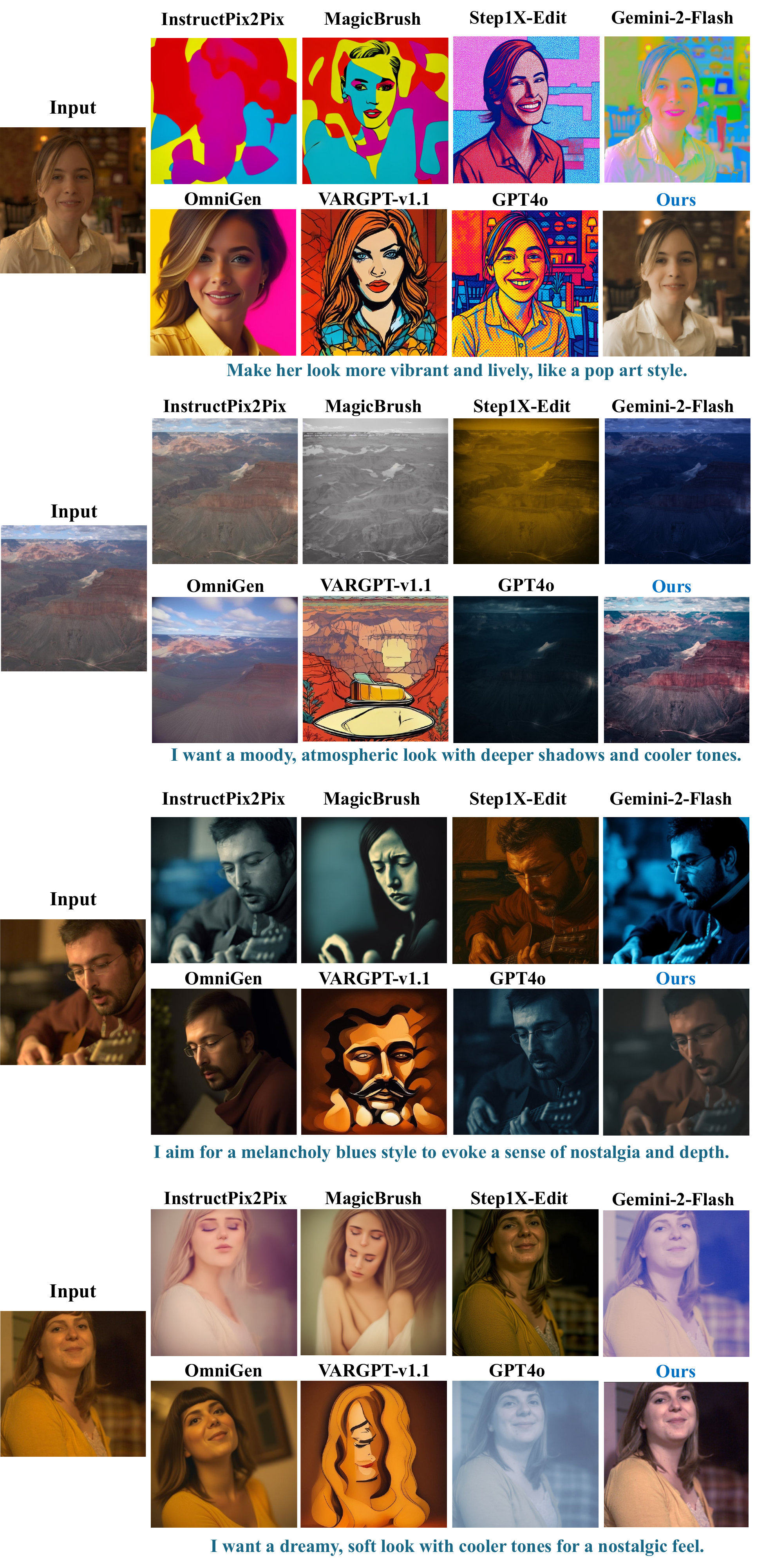}
  \caption{Visual comparisons of all instruction-based editing methods on MIT-FiveK~\cite{fivek}.}
  \label{fig:five5k_results1}
\end{figure*}

\begin{figure*}[!t]
  \vspace{-1.5cm}
  \setlength{\abovecaptionskip}{-0.1cm} 
  \centering
  \includegraphics[width=0.85\linewidth]{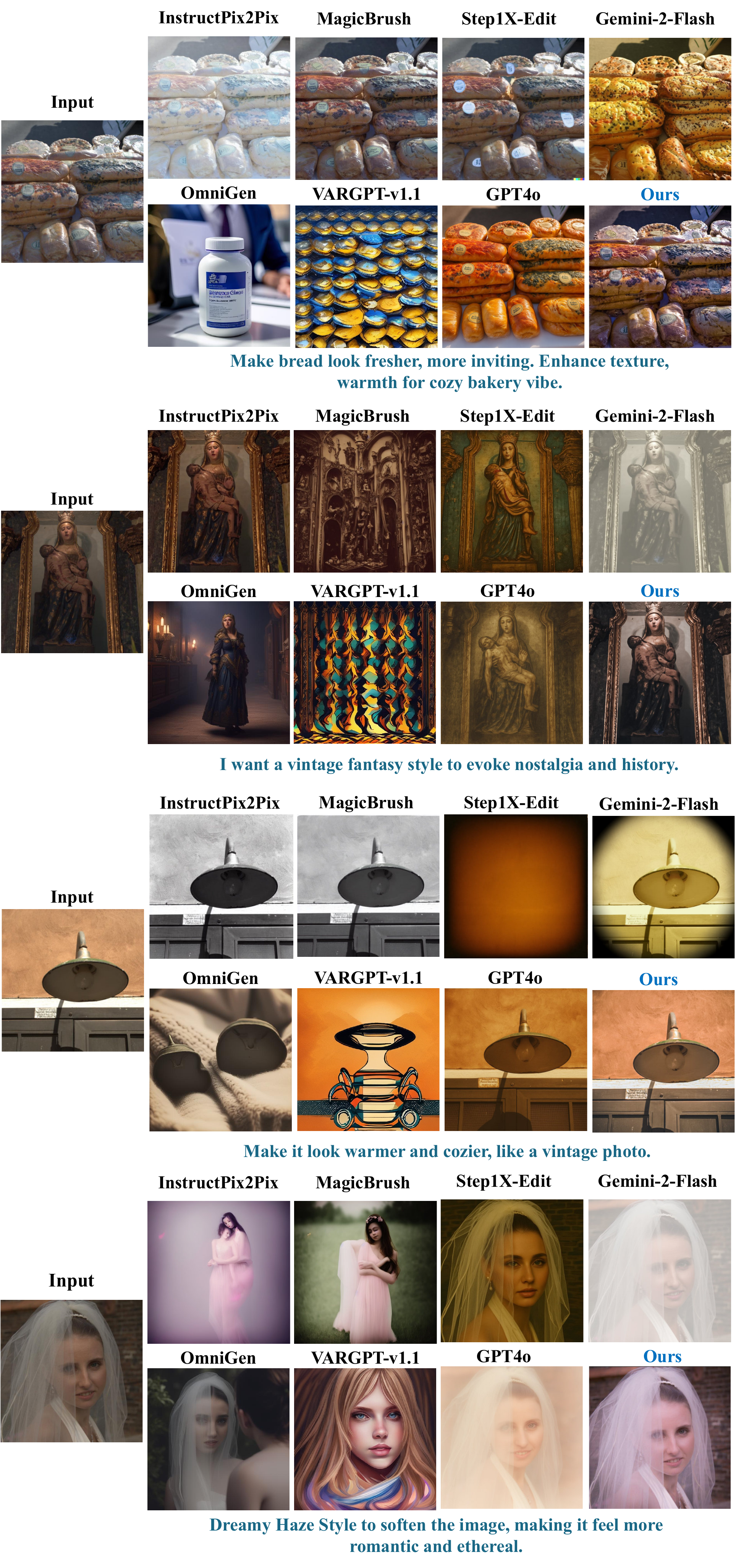}
  \caption{Visual comparisons of all instruction-based editing methods on MIT-FiveK~\cite{fivek}.}
  \label{fig:five5k_results2}
\end{figure*}

To assess the generalization ability of our system, we conduct comprehensive qualitative and visual comparisons on the MIT-FiveK~\cite{fivek} benchmark dataset. Specifically, we randomly select 50 data samples from MIT-FiveK~\cite{fivek} and generate corresponding user instructions based on the source images, expert C's reference images, and the associated retouching parameter configurations (see Section 3.2 for implementation details). As shown in Table~\ref{tab:five5K} and Figures~\ref{fig:five5k_results1} and~\ref{fig:five5k_results2}, our system achieves state-of-the-art performance in both instruction-following and content preservation metrics. These results confirm the robustness and effectiveness of our method in faithfully executing user instructions while maintaining original image content. Furthermore, our approach consistently outperforms existing baselines on multiple real-world benchmarks, highlighting its strong generalization ability and practical applicability in interactive photo retouching.

\begin{figure*}[htbp]
    \centering
    \includegraphics[width=\linewidth]{figs_Appendix/prompt_metrics.pdf}
    \caption{Prompt for MLLM-based metrics (SC, PQ) from scene-level and region-level.
    }
    \label{prompt_metrics}
\end{figure*}

\begin{figure*}[htbp]
    \vspace{-2cm}
    \centering
    \includegraphics[width=\linewidth]{figs_Appendix/prompt_preset_recom.pdf}
    \caption{Role-playing prompt for preset recommendation.
    }
    \label{prompt_preset_recom}
\end{figure*}

\begin{figure*}[htbp]
    \vspace{-2cm}
    \centering
    \includegraphics[width=\linewidth]{figs_Appendix/prompt_user_casual.pdf}
    \caption{Prompt for simulating the casual user instructions.
    }
    \label{prompt_user_casual}
\end{figure*}

\clearpage
\newpage
\section{Details of Retouching Tools in Lightroom}\label{app:tools_detials}
We provide an overview of key Lightroom tools generated by JarvisArt, focusing on the functionality of retouching tools and their associated parameters:

\begin{longtable}{>{\RaggedRight}p{0.3\textwidth} >{\RaggedRight}p{0.52\textwidth} >{\RaggedRight\arraybackslash}p{0.1\textwidth}}
\caption{Lightroom Tools with Functional Description and Parameter Type.}\label{tab:model_params}\\
\toprule
\textbf{Tool Name} & \textbf{Functional Description } & \textbf{Type} \\
\midrule
\endfirsthead
\caption[]{Lightroom tools with functional description and parameter type. (Continued)}\\
\toprule
\textbf{Tool Name} & \textbf{Functional Description} & \textbf{Type} \\
\midrule
\endhead
\midrule
\multicolumn{3}{r}{\textit{Continued on next page}} \\
\endfoot
\bottomrule
\endlastfoot

\multicolumn{3}{l}{\param{\textbf{Basic Adjustments}}} \\
WhiteBalance & Overall color temperature (As Shot, Auto, Custom) & Str. \\
Temperature & Blue-yellow balance (2000-10000 Kelvin) & Num. \\
Tint & Green-magenta balance (-150 to +150) & Num. \\
Exposure2012 & Overall brightness (-5.0 to +5.0 stops) & Num. \\
Contrast2012 & Difference between light/dark areas (-100 to +100) & Num. \\
Highlights2012 & Adjusts bright areas (-100 to +100) & Num. \\
Shadows2012 & Adjusts dark areas (-100 to +100) & Num. \\
Whites2012 & Fine-tunes brightest parts (-100 to +100) & Num. \\
Blacks2012 & Fine-tunes darkest parts (-100 to +100) & Num. \\
Texture & Enhances/smooths medium textures (-100 to +100) & Num. \\
Clarity2012 & Enhances/reduces local mid-tone contrast (-100 to +100) & Num. \\
Dehaze & Reduces/adds atmospheric haze (-100 to +100) & Num. \\
Vibrance & Saturation of less-saturated colors (-100 to +100) & Num. \\
Saturation & Overall color intensity (-100 to +100) & Num. \\
IncrementalTemperature & Relative temperature adjustment (-100 to +100) & Num. \\
IncrementalTint & Relative tint adjustment (-100 to +100) & Num. \\
\midrule

\multicolumn{3}{l}{\param{\textbf{Tone Curve}}} \\
ToneCurveName2012 & Predefined curve shape (Linear, Custom) & Str. \\
ToneCurvePV2012 & Custom RGB tone curve points (x,y: 0-255) & Dict. \\
ToneCurvePV2012Red & Custom Red channel tone curve points & Dict. \\
ToneCurvePV2012Green & Custom Green channel tone curve points & Dict. \\
ToneCurvePV2012Blue & Custom Blue channel tone curve points & Dict. \\
ParametricShadows & Adjusts shadow tonal regions (-100 to +100) & Num. \\
ParametricDarks & Adjusts dark tonal regions (-100 to +100) & Num. \\
ParametricLights & Adjusts light tonal regions (-100 to +100) & Num. \\
ParametricHighlights & Adjusts highlight tonal regions (-100 to +100) & Num. \\
ParametricShadowSplit & Boundary: shadows/darks (10-50) & Num. \\
ParametricMidtoneSplit & Boundary: darks/lights (25-75) & Num. \\
ParametricHighlightSplit & Boundary: lights/highlights (50-90) & Num. \\
\midrule

\multicolumn{3}{l}{\param{\textbf{Detail}}} \\
Sharpness & Enhances edge definition (0-150) & Num. \\
SharpenRadius & Width of sharpening effect (0.5-3.0) & Num. \\
SharpenDetail & Amount of sharpening for details (0-100) & Num. \\
SharpenEdgeMasking & Masks sharpening to edges (0-100) & Num. \\
LuminanceSmoothing & Reduces luminance noise (0-100) & Num. \\
ColorNoiseReduction & Reduces color noise (0-100) & Num. \\
ColorNoiseReductionDetail & Fine-tunes color noise reduction (0-100) & Num. \\
ColorNoiseReductionSmoothness & Smoothness of color noise reduction (0-100) & Num. \\
\midrule

\multicolumn{3}{l}{\param{\textbf{HSL/Color (per color: Red, Orange, Yellow, Green, Aqua, Blue, Purple, Magenta)}}} \\
HueAdjustment<Color> & Shifts hue of specific color (-100 to +100) & Num. \\
SaturationAdjustment<Color> & Adjusts saturation of specific color (-100 to +100) & Num. \\
LuminanceAdjustment<Color> & Adjusts brightness of specific color (-100 to +100) & Num. \\
\midrule

\multicolumn{3}{l}{\param{\textbf{Color Grading}}} \\
SplitToningShadowHue & Hue for shadows in split toning (0-359) & Num. \\
SplitToningHighlightHue & Hue for highlights in split toning (0-359) & Num. \\
SplitToningShadowSaturation & Saturation for shadows (0-100) & Num. \\
SplitToningHighlightSaturation & Saturation for highlights (0-100) & Num. \\
SplitToningBalance & Balance between shadow/highlight toning (-100 to +100) & Num. \\
ColorGradeMidtoneHue & Midtone hue for color grading (0-359) & Num. \\
ColorGradeMidtoneSat & Midtone saturation for color grading (0-100) & Num. \\
ColorGradeMidtoneLum & Midtone luminance for color grading (0-100) & Num. \\
ColorGradeShadowLum & Luminance for shadows (0-100) & Num. \\
ColorGradeHighlightLum & Luminance for highlights (0-100) & Num. \\
ColorGradeBlending & Blending of color grading effect (0-100) & Num. \\
ColorGradeGlobalHue & Global hue adjustment (0-359) & Num. \\
ColorGradeGlobalSat & Global saturation adjustment (0-100) & Num. \\
ColorGradeGlobalLum & Global luminance adjustment (0-100) & Num. \\
\midrule

\multicolumn{3}{l}{\param{\textbf{Effects}}} \\
PostCropVignetteAmount & Darkens/lightens image corners (-100 to +100) & Num. \\
GrainAmount & Adds film grain effect (0-100) & Num. \\
ShadowTint & Adjusts color tint in shadows (-100 to +100) & Num. \\
\midrule

\multicolumn{3}{l}{\param{\textbf{Camera Calibration (for Red, Green, Blue primary channels)}}} \\
<PrimaryColor>Hue & Shifts primary color's hue (-100 to +100) & Num. \\
<PrimaryColor>Saturation & Adjusts primary color's saturation (-100 to +100) & Num. \\
\midrule

\multicolumn{3}{l}{\param{\textbf{Lens Blur (Overall: Dict.)}}} \\
LensBlur.Active & Enables/disables lens blur effect & Bool. \\
LensBlur.BlurAmount & Strength of blur effect (0-100) & Num. \\
LensBlur.FocalRange & Defines focal plane ("x1 y1 x2 y2") & Str. \\
LensBlur.BokehShape & Bokeh shape identifier (default 0) & Num. \\
LensBlur.BokehShapeDetail & Definition of bokeh shape edges (0-100) & Num. \\
LensBlur.HighlightsThreshold & Brightness threshold for bokeh (0-100) & Num. \\
LensBlur.HighlightsBoost & Enhances out-of-focus highlights (0-100) & Num. \\
LensBlur.CatEyeAmount & Simulates cat's eye bokeh effect (0-100) & Num. \\
LensBlur.CatEyeScale & Size of cat's eye effect (0-100) & Num. \\
\midrule

\multicolumn{3}{l}{\param{\textbf{Advanced Color Grading (PointColors - each point is a Dict.)}}} \\
SrcHue & Source hue for adjustment (0-6.28 rad) & Num. \\
SrcSat & Source saturation for adjustment (0-1.0) & Num. \\
SrcLum & Source luminance for adjustment (0-1.0) & Num. \\
HueShift & Hue shift amount (-1 to +1) & Num. \\
SatScale & Saturation scale (-1 to +1) & Num. \\
LumScale & Luminance scale (-1 to +1) & Num. \\
RangeAmount & Effect application amount (0-1.0) & Num. \\
HueRange & Falloff for hue adjustment (LowerNone, LowerFull, UpperFull, UpperNone: 0-1.0) & Dict. \\
SatRange & Falloff for saturation adjustment (sub-props same as HueRange) & Dict. \\
LumRange & Falloff for luminance adjustment (sub-props same as HueRange) & Dict. \\
\midrule

\multicolumn{3}{l}{\param{\textbf{Look (Overall: Dict.)}}} \\
Look.Name & Name of the look preset & Str. \\
Look.Amount & Intensity of the look effect (0.0-1.0) & Num. \\
Look.Parameters & Dictionary of specific adjustments applied by the look & Dict. \\
\ \ \textit{ (e.g., ProcessVersion, ToneCurvePV2012, Parametric adjustments, SplitToning, ColorGrade, ConvertToGrayscale, LookTable, RGBTable, RGBTableAmount)} & & \\
\midrule

\multicolumn{3}{l}{\param{\textbf{Localized Mask Adjustments (MaskGroupBasedCorrections - Array of Dicts.)}}} \\
\textit{Per Correction Group:} \\
CorrectionAmount & Amount for the correction group (0-1, default 1) & Num. \\
CorrectionActive & Activates the correction group & Bool. \\
CorrectionName & Name for the correction group & Str. \\
LocalExposure2012 & Local exposure adjustment (-1 to +1) & Num. \\
LocalContrast2012 & Local contrast adjustment (-1 to +1) & Num. \\
LocalHighlights2012 & Local highlights adjustment (-1 to +1) & Num. \\
LocalShadows2012 & Local shadows adjustment (-1 to +1) & Num. \\
LocalWhites2012 & Local whites adjustment (-1 to +1) & Num. \\
LocalBlacks2012 & Local blacks adjustment (-1 to +1) & Num. \\
LocalClarity / LocalClarity2012 & Local clarity adjustment (-1 to +1) & Num. \\
LocalDehaze & Local dehaze adjustment (-1 to +1) & Num. \\
LocalTexture & Local texture adjustment (-1 to +1) & Num. \\
LocalHue & Local hue adjustment (-1 to +1) & Num. \\
LocalSaturation & Local saturation adjustment (-1 to +1) & Num. \\
LocalCurveRefineSaturation & Local saturation curve refinement (0-100) & Num. \\
LocalToningHue & Local toning hue (0-359) & Num. \\
LocalToningSaturation & Local toning saturation (-1 to +1) & Num. \\
LocalTemperature & Local temperature adjustment (-1 to +1) & Num. \\
LocalTint & Local tint adjustment (-1 to +1) & Num. \\
LocalLuminanceNoise & Local luminance noise reduction (-1 to +1) & Num. \\
LocalMoire & Local moire reduction (-1 to +1) & Num. \\
LocalDefringe & Local defringe adjustment (-1 to +1) & Num. \\
LocalGrain & Local grain adjustment (-1 to +1) & Num. \\
LocalSharpness & Local sharpness adjustment (-1 to +1) & Num. \\
<Channel>Curve & Local tone curve for Red, Green, Blue, or Main channels (points "x,y") & Dict. \\
LocalPointColors & Local specific color adjustments (dictionary of string-encoded points) & Dict. \\
CorrectionMasks & Array of mask definitions for the group & Array \\
\ \ \textit{Per Mask in CorrectionMasks:} \\
\ \ What & Mask type (e.g., "Mask/Image", "Mask/CircularGradient") & Str. \\
\ \ MaskActive & Activates this specific mask & Bool. \\
\ \ MaskName & Name of the mask (e.g., "Subject", "Sky") & Str. \\
\ \ MaskBlendMode & Mask blending (0=Add, 1=Intersect) & Num. \\
\ \ MaskInverted & Inverts the mask area & Bool. \\
\ \ MaskValue & Mask opacity (0.0-1.0) & Num. \\
\ \ MaskSubType & AI Mask subtype (Subject, Sky, Person etc.) / Object type & Num. \\
\ \ ReferencePoint & Center point for AI masks ("x y") & Str. \\
\ \ Gesture & Polygon points for object/region mask & Array \\
\ \ Top/Left/Bottom/Right & Coordinates for radial gradient (0-1) & Num. \\
\ \ Angle & Rotation angle for radial gradient (0-360) & Num. \\
\ \ Midpoint & Center point of radial gradient (0-100) & Num. \\
\ \ Feather & Edge feathering for radial gradient (0-100) & Num. \\
\ \ Flipped & Flips radial gradient direction & Bool. \\
\ \ MaskSubCategoryID & Category ID for person parts mask (Face, Eyes, etc.) & Num. \\

\end{longtable}

\clearpage
\newpage
\small{
\normalem
\bibliographystyle{abbrv}
\bibliography{ref}

\begin{thebibliography}{10}

\bibitem{bai2025qwen2}
S.~Bai, K.~Chen, X.~Liu, J.~Wang, W.~Ge, S.~Song, K.~Dang, P.~Wang, S.~Wang, J.~Tang, et~al.
\newblock Qwen2. 5-vl technical report.
\newblock {\em arXiv preprint arXiv:2502.13923}, 2025.

\bibitem{brooks2023instructpix2pix}
T.~Brooks, A.~Holynski, and A.~A. Efros.
\newblock Instructpix2pix: Learning to follow image editing instructions.
\newblock In {\em Proceedings of the IEEE/CVF conference on computer vision and pattern recognition}, pages 18392--18402, 2023.

\bibitem{fivek}
V.~Bychkovsky, S.~Paris, E.~Chan, and F.~Durand.
\newblock Learning photographic global tonal adjustment with a database of input / output image pairs.
\newblock In {\em The Twenty-Fourth IEEE Conference on Computer Vision and Pattern Recognition}, 2011.

\bibitem{Cai2024Internlm2}
Z.~Cai, M.~Cao, H.~Chen, K.~Chen, K.~Chen, X.~Chen, X.~Chen, Z.~Chen, Z.~Chen, P.~Chu, et~al.
\newblock Internlm2 technical report.
\newblock {\em arXiv preprint arXiv:2403.17297}, 2024.

\bibitem{chen2025r1v}
L.~Chen, L.~Li, H.~Zhao, Y.~Song, and Vinci.
\newblock R1-v: Reinforcing super generalization ability in vision-language models with less than \$3.
\newblock {\em GitHub repository: https://github.com/Deep-Agent/R1-V}, 2025.

\bibitem{chen2017zoo}
P.-Y. Chen, H.~Zhang, Y.~Sharma, J.~Yi, and C.-J. Hsieh.
\newblock Zoo: Zeroth order optimization based black-box attacks to deep neural networks without training substitute models.
\newblock In {\em Proceedings of the 10th ACM workshop on artificial intelligence and security}, pages 15--26, 2017.

\bibitem{chen2025janus}
X.~Chen, Z.~Wu, X.~Liu, Z.~Pan, W.~Liu, Z.~Xie, X.~Yu, and C.~Ruan.
\newblock Janus-pro: Unified multimodal understanding and generation with data and model scaling.
\newblock {\em arXiv preprint arXiv:2501.17811}, 2025.

\bibitem{fu2024mgie}
T.-J. Fu, W.~Hu, X.~Du, W.~Y. Wang, Y.~Yang, and Z.~Gan.
\newblock Guiding instruction-based image editing via multimodal large language models.
\newblock In {\em International Conference on Learning Representations (ICLR)}, 2024.

\bibitem{guo2025deepseek}
D.~Guo, D.~Yang, H.~Zhang, J.~Song, R.~Zhang, R.~Xu, Q.~Zhu, S.~Ma, P.~Wang, X.~Bi, et~al.
\newblock Deepseek-r1: Incentivizing reasoning capability in llms via reinforcement learning.
\newblock {\em arXiv preprint arXiv:2501.12948}, 2025.

\bibitem{guo2024lightrag}
Z.~Guo, L.~Xia, Y.~Yu, T.~Ao, and C.~Huang.
\newblock Lightrag: Simple and fast retrieval-augmented generation.
\newblock 2024.

\bibitem{hansen2006cma}
N.~Hansen.
\newblock The cma evolution strategy: a comparing review.
\newblock {\em Towards a new evolutionary computation: Advances in the estimation of distribution algorithms}, pages 75--102, 2006.

\bibitem{hong2023metagpt}
S.~Hong, X.~Zheng, J.~Chen, Y.~Cheng, J.~Wang, C.~Zhang, Z.~Wang, S.~K.~S. Yau, Z.~Lin, L.~Zhou, et~al.
\newblock Metagpt: Meta programming for multi-agent collaborative framework.
\newblock {\em arXiv preprint arXiv:2308.00352}, 3(4):6, 2023.

\bibitem{hu2018exposure}
Y.~Hu, H.~He, C.~Xu, B.~Wang, and S.~Lin.
\newblock Exposure: A white-box photo post-processing framework.
\newblock {\em ACM Transactions on Graphics (TOG)}, 37(2):1--17, 2018.

\bibitem{huang2024audiogpt}
R.~Huang, M.~Li, D.~Yang, J.~Shi, X.~Chang, Z.~Ye, Y.~Wu, Z.~Hong, J.~Huang, J.~Liu, et~al.
\newblock Audiogpt: Understanding and generating speech, music, sound, and talking head.
\newblock In {\em Proceedings of the AAAI Conference on Artificial Intelligence}, pages 23802--23804, 2024.

\bibitem{huang2025visionr1}
W.~Huang, B.~Jia, Z.~Zhai, S.~Cao, Z.~Ye, F.~Zhao, Y.~Hu, and S.~Lin.
\newblock Vision-r1: Incentivizing reasoning capability in multimodal large language models.
\newblock {\em arXiv preprint arXiv:2503.06749}, 2025.

\bibitem{Hui2024Qwen25coder}
B.~Hui, J.~Yang, Z.~Cui, J.~Yang, D.~Liu, L.~Zhang, T.~Liu, J.~Zhang, B.~Yu, K.~Lu, et~al.
\newblock Qwen2.5-coder technical report.
\newblock {\em arXiv preprint arXiv:2409.12186}, 2024.

\bibitem{hui2024hq}
M.~Hui, S.~Yang, B.~Zhao, Y.~Shi, H.~Wang, P.~Wang, Y.~Zhou, and C.~Xie.
\newblock Hq-edit: A high-quality dataset for instruction-based image editing.
\newblock {\em arXiv preprint arXiv:2404.09990}, 2024.

\bibitem{hurst2024gpt}
A.~Hurst, A.~Lerer, A.~P. Goucher, A.~Perelman, A.~Ramesh, A.~Clark, A.~Ostrow, A.~Welihinda, A.~Hayes, A.~Radford, et~al.
\newblock Gpt-4o system card.
\newblock {\em arXiv preprint arXiv:2410.21276}, 2024.

\bibitem{Jaech2024Openai}
A.~Jaech, A.~Kalai, A.~Lerer, A.~Richardson, A.~El-Kishky, A.~Low, A.~Helyar, A.~Madry, A.~Beutel, A.~Carney, et~al.
\newblock Openai o1 system card.
\newblock {\em arXiv preprint arXiv:2412.16720}, 2024.

\bibitem{Jiao2024Preference}
F.~Jiao, G.~Guo, X.~Zhang, N.~F. Chen, S.~Joty, and F.~Wei.
\newblock Preference optimization for reasoning with pseudo feedback.
\newblock {\em arXiv preprint arXiv:2411.16345}, 2024.

\bibitem{jin2025search}
B.~Jin, H.~Zeng, Z.~Yue, D.~Wang, H.~Zamani, and J.~Han.
\newblock Search-r1: Training llms to reason and leverage search engines with reinforcement learning.
\newblock {\em arXiv preprint arXiv:2503.09516}, 2025.

\bibitem{ke2022harmonizer}
Z.~Ke, C.~Sun, L.~Zhu, K.~Xu, and R.~W. Lau.
\newblock Harmonizer: Learning to perform white-box image and video harmonization.
\newblock In {\em European conference on computer vision}, pages 690--706. Springer, 2022.

\bibitem{kosugi2020unpaired}
S.~Kosugi and T.~Yamasaki.
\newblock Unpaired image enhancement featuring reinforcement-learning-controlled image editing software.
\newblock In {\em Proceedings of the AAAI conference on artificial intelligence}, pages 11296--11303, 2020.

\bibitem{ku2023viescore}
M.~Ku, D.~Jiang, C.~Wei, X.~Yue, and W.~Chen.
\newblock Viescore: Towards explainable metrics for conditional image synthesis evaluation.
\newblock {\em arXiv preprint arXiv:2312.14867}, 2023.

\bibitem{langchain}
LangChain.
\newblock Langchain: Build context-aware reasoning applications.
\newblock \url{https://github.com/langchain-ai/langchain}, 2023.

\bibitem{liang2021ppr10k}
J.~Liang, H.~Zeng, M.~Cui, X.~Xie, and L.~Zhang.
\newblock Ppr10k: A large-scale portrait photo retouching dataset with human-region mask and group-level consistency.
\newblock In {\em Proceedings of the IEEE/CVF Conference on Computer Vision and Pattern Recognition}, pages 653--661, 2021.

\bibitem{jarvisir2025}
Y.~Lin, Z.~Lin, H.~Chen, P.~Pan, C.~Li, S.~Chen, W.~Kairun, Y.~Jin, W.~Li, and X.~Ding.
\newblock Jarvisir: Elevating autonomous driving perception with intelligent image restoration.
\newblock In {\em Proceedings of the IEEE/CVF Conference on Computer Vision and Pattern Recognition (CVPR)}, 2025.

\bibitem{liu2025step1x}
S.~Liu, Y.~Han, P.~Xing, F.~Yin, R.~Wang, W.~Cheng, J.~Liao, Y.~Wang, H.~Fu, C.~Han, et~al.
\newblock Step1x-edit: A practical framework for general image editing.
\newblock {\em arXiv preprint arXiv:2504.17761}, 2025.

\bibitem{liu2024grounding}
S.~Liu, Z.~Zeng, T.~Ren, F.~Li, H.~Zhang, J.~Yang, Q.~Jiang, C.~Li, J.~Yang, H.~Su, et~al.
\newblock Grounding dino: Marrying dino with grounded pre-training for open-set object detection.
\newblock In {\em European Conference on Computer Vision}, pages 38--55. Springer, 2024.

\bibitem{liu2024apigen}
Z.~Liu, T.~Hoang, J.~Zhang, M.~Zhu, T.~Lan, J.~Tan, W.~Yao, Z.~Liu, Y.~Feng, R.~RN, et~al.
\newblock Apigen: Automated pipeline for generating verifiable and diverse function-calling datasets.
\newblock {\em Advances in Neural Information Processing Systems}, 37:54463--54482, 2024.

\bibitem{liu2025visualrft}
Z.~Liu, Z.~Sun, Y.~Zang, X.~Dong, Y.~Cao, H.~Duan, D.~Lin, and J.~Wang.
\newblock Visual-rft: Visual reinforcement fine-tuning.
\newblock {\em arXiv preprint arXiv:2503.01785}, 2025.

\bibitem{lu2025uir1}
Z.~Lu, Y.~Chai, Y.~Guo, X.~Yin, L.~Liu, H.~Wang, G.~Xiong, and H.~Li.
\newblock Ui-r1: Enhancing action prediction of gui agents by reinforcement learning.
\newblock {\em arXiv preprint arXiv:2503.21620}, 2025.

\bibitem{ma2025unitok}
C.~Ma, Y.~Jiang, J.~Wu, J.~Yang, X.~Yu, Z.~Yuan, B.~Peng, and X.~Qi.
\newblock Unitok: A unified tokenizer for visual generation and understanding.
\newblock {\em arXiv preprint arXiv:2502.20321}, 2025.

\bibitem{meng2025mmeureka}
F.~Meng, L.~Du, Z.~Liu, Z.~Zhou, Q.~Lu, D.~Fu, B.~Shi, W.~Wang, J.~He, K.~Zhang, et~al.
\newblock Mm-eureka: Exploring visual aha moment with rule-based large-scale reinforcement learning.
\newblock {\em arXiv preprint arXiv:2503.07365}, 2025.

\bibitem{mosleh2020hardware}
A.~Mosleh, A.~Sharma, E.~Onzon, F.~Mannan, N.~Robidoux, and F.~Heide.
\newblock Hardware-in-the-loop end-to-end optimization of camera image processing pipelines.
\newblock In {\em Proceedings of the IEEE/CVF Conference on Computer Vision and Pattern Recognition}, pages 7529--7538, 2020.

\bibitem{nishimura2018automatic}
J.~Nishimura, T.~Gerasimow, R.~Sushma, A.~Sutic, C.-T. Wu, and G.~Michael.
\newblock Automatic isp image quality tuning using nonlinear optimization.
\newblock In {\em 2018 25th IEEE International Conference on Image Processing (ICIP)}, pages 2471--2475. IEEE, 2018.

\bibitem{ouyang2023rsfnet}
W.~Ouyang, Y.~Dong, X.~Kang, P.~Ren, X.~Xu, and X.~Xie.
\newblock Rsfnet: A white-box image retouching approach using region-specific color filters.
\newblock In {\em Proceedings of the IEEE/CVF International Conference on Computer Vision}, pages 12160--12169, 2023.

\bibitem{pan2025metaspatial}
Z.~Pan and H.~Liu.
\newblock Metaspatial: Reinforcing 3d spatial reasoning in vlms for the metaverse.
\newblock {\em arXiv preprint arXiv:2503.18470}, 2025.

\bibitem{qian2025toolrl}
C.~Qian, E.~C. Acikgoz, Q.~He, H.~Wang, X.~Chen, D.~Hakkani-T{\"u}r, G.~Tur, and H.~Ji.
\newblock Toolrl: Reward is all tool learning needs.
\newblock {\em arXiv preprint arXiv:2504.13958}, 2025.

\bibitem{shao2024deepseekmath}
Z.~Shao, P.~Wang, Q.~Zhu, R.~Xu, J.~Song, X.~Bi, H.~Zhang, M.~Zhang, Y.~Li, Y.~Wu, et~al.
\newblock Deepseekmath: Pushing the limits of mathematical reasoning in open language models.
\newblock {\em arXiv preprint arXiv:2402.03300}, 2024.

\bibitem{shen2025vlm}
H.~Shen, P.~Liu, J.~Li, C.~Fang, Y.~Ma, J.~Liao, Q.~Shen, Z.~Zhang, K.~Zhao, Q.~Zhang, et~al.
\newblock Vlm-r1: A stable and generalizable r1-style large vision-language model.
\newblock {\em arXiv preprint arXiv:2504.07615}, 2025.

\bibitem{sheng2024hybridflow}
G.~Sheng, C.~Zhang, Z.~Ye, X.~Wu, W.~Zhang, R.~Zhang, Y.~Peng, H.~Lin, and C.~Wu.
\newblock Hybridflow: A flexible and efficient rlhf framework.
\newblock {\em arXiv preprint arXiv: 2409.19256}, 2024.

\bibitem{autogpt}
Significant-Gravitas.
\newblock Autogpt.
\newblock \url{https://github.com/Significant-Gravitas/AutoGPT}, 2023.

\bibitem{team2023gemini}
G.~Team, R.~Anil, S.~Borgeaud, J.-B. Alayrac, J.~Yu, R.~Soricut, J.~Schalkwyk, A.~M. Dai, A.~Hauth, K.~Millican, et~al.
\newblock Gemini: a family of highly capable multimodal models.
\newblock {\em arXiv preprint arXiv:2312.11805}, 2023.

\bibitem{tseng2019hyperparameter}
E.~Tseng, F.~Yu, Y.~Yang, F.~Mannan, K.~S. Arnaud, D.~Nowrouzezahrai, J.-F. Lalonde, and F.~Heide.
\newblock Hyperparameter optimization in black-box image processing using differentiable proxies.
\newblock {\em ACM Trans. Graph.}, 38(4):27--1, 2019.

\bibitem{tseng2022neural}
E.~Tseng, Y.~Zhang, L.~Jebe, X.~Zhang, Z.~Xia, Y.~Fan, F.~Heide, and J.~Chen.
\newblock Neural photo-finishing.
\newblock {\em ACM Trans. Graph.}, 41(6):238--1, 2022.

\bibitem{wu2024goal}
J.~Wu, Y.~Wang, L.~Li, F.~Zhang, and T.~Xue.
\newblock Goal conditioned reinforcement learning for photo finishing tuning.
\newblock {\em Advances in Neural Information Processing Systems}, 37:46294--46318, 2024.

\bibitem{xia2025gui}
X.~Xia and R.~Luo.
\newblock Gui-r1: A generalist r1-style vision-language action model for gui agents.
\newblock {\em arXiv preprint arXiv:2504.10458}, 2025.

\bibitem{xiao2024omnigen}
S.~Xiao, Y.~Wang, J.~Zhou, H.~Yuan, X.~Xing, R.~Yan, S.~Wang, T.~Huang, and Z.~Liu.
\newblock Omnigen: Unified image generation.
\newblock {\em arXiv preprint arXiv:2409.11340}, 2024.

\bibitem{xie2023openagents}
T.~Xie, F.~Zhou, Z.~Cheng, P.~Shi, L.~Weng, Y.~Liu, T.~J. Hua, J.~Zhao, Q.~Liu, C.~Liu, et~al.
\newblock Openagents: An open platform for language agents in the wild.
\newblock {\em arXiv preprint arXiv:2310.10634}, 2023.

\bibitem{yang2024qwen2}
A.~Yang, B.~Yang, B.~Zhang, B.~Hui, B.~Zheng, B.~Yu, C.~Li, D.~Liu, F.~Huang, H.~Wei, et~al.
\newblock Qwen2.5 technical report.
\newblock {\em arXiv preprint arXiv:2412.15115}, 2024.

\bibitem{Yang2024Qwen25math}
A.~Yang, B.~Zhang, B.~Hui, B.~Gao, B.~Yu, C.~Li, D.~Liu, J.~Tu, J.~Zhou, J.~Lin, et~al.
\newblock Qwen2.5-math technical report: Toward mathematical expert model via self-improvement.
\newblock {\em arXiv preprint arXiv:2409.12122}, 2024.

\bibitem{yang2025r1}
Y.~Yang, X.~He, H.~Pan, X.~Jiang, Y.~Deng, X.~Yang, H.~Lu, D.~Yin, F.~Rao, M.~Zhu, et~al.
\newblock R1-onevision: Advancing generalized multimodal reasoning through cross-modal formalization.
\newblock {\em arXiv preprint arXiv:2503.10615}, 2025.

\bibitem{yao2023react}
S.~Yao, J.~Zhao, D.~Yu, N.~Du, I.~Shafran, K.~Narasimhan, and Y.~Cao.
\newblock React: Synergizing reasoning and acting in language models.
\newblock In {\em International Conference on Learning Representations (ICLR)}, 2023.

\bibitem{Ying2024Internlm}
H.~Ying, S.~Zhang, L.~Li, Z.~Zhou, Y.~Shao, Z.~Fei, Y.~Ma, J.~Hong, K.~Liu, Z.~Wang, et~al.
\newblock Internlm-math: Open math large language models toward verifiable reasoning.
\newblock {\em arXiv preprint arXiv:2402.06332}, 2024.

\bibitem{yu2021reconfigisp}
K.~Yu, Z.~Li, Y.~Peng, C.~C. Loy, and J.~Gu.
\newblock Reconfigisp: Reconfigurable camera image processing pipeline.
\newblock In {\em Proceedings of the IEEE/CVF International Conference on Computer Vision}, pages 4248--4257, 2021.

\bibitem{zeng2020learning}
H.~Zeng, J.~Cai, L.~Li, Z.~Cao, and L.~Zhang.
\newblock Learning image-adaptive 3d lookup tables for high performance photo enhancement in real-time.
\newblock {\em IEEE Transactions on Pattern Analysis and Machine Intelligence}, 44(4):2058--2073, 2020.

\bibitem{Zhang2024Codedpo}
K.~Zhang, G.~Li, Y.~Dong, J.~Xu, J.~Zhang, J.~Su, Y.~Liu, and Z.~Jin.
\newblock Codedpo: Aligning code models with self generated and verified source code.
\newblock {\em arXiv preprint arXiv:2410.05605}, 2024.

\bibitem{zhang2023magicbrush}
K.~Zhang, L.~Mo, W.~Chen, H.~Sun, and Y.~Su.
\newblock Magicbrush: A manually annotated dataset for instruction-guided image editing.
\newblock {\em Advances in Neural Information Processing Systems}, 36:31428--31449, 2023.

\bibitem{zhang1996spatial}
X.~Zhang, B.~A. Wandell, et~al.
\newblock A spatial extension of cielab for digital color image reproduction.
\newblock In {\em SID international symposium digest of technical papers}, volume~27, pages 731--734. Citeseer, 1996.

\bibitem{Zhang2024o1coder}
Y.~Zhang, S.~Wu, Y.~Yang, J.~Shu, J.~Xiao, C.~Kong, and J.~Sang.
\newblock o1-coder: an o1 replication for coding.
\newblock {\em arXiv preprint arXiv:2412.00154}, 2024.

\bibitem{zhang2024survey}
Z.~Zhang, X.~Bo, C.~Ma, R.~Li, X.~Chen, Q.~Dai, J.~Zhu, Z.~Dong, and J.-R. Wen.
\newblock A survey on the memory mechanism of large language model based agents.
\newblock {\em arXiv preprint arXiv:2404.13501}, 2024.

\bibitem{zhao2024ultraedit}
H.~Zhao, X.~S. Ma, L.~Chen, S.~Si, R.~Wu, K.~An, P.~Yu, M.~Zhang, Q.~Li, and B.~Chang.
\newblock Ultraedit: Instruction-based fine-grained image editing at scale.
\newblock {\em Advances in Neural Information Processing Systems}, 37:3058--3093, 2024.

\bibitem{zhao2025qlip}
Y.~Zhao, F.~Xue, S.~Reed, L.~Fan, Y.~Zhu, J.~Kautz, Z.~Yu, P.~Kr{\"a}henb{\"u}hl, and D.-A. Huang.
\newblock Qlip: Text-aligned visual tokenization unifies auto-regressive multimodal understanding and generation.
\newblock {\em arXiv preprint arXiv:2502.05178}, 2025.

\bibitem{zheng2024llamafactory}
Y.~Zheng, R.~Zhang, J.~Zhang, Y.~Ye, Z.~Luo, Z.~Feng, and Y.~Ma.
\newblock Llamafactory: Unified efficient fine-tuning of 100+ language models.
\newblock In {\em Proceedings of the 62nd Annual Meeting of the Association for Computational Linguistics (Volume 3: System Demonstrations)}, Bangkok, Thailand, 2024. Association for Computational Linguistics.

\bibitem{zhuang2025vargpt}
X.~Zhuang, Y.~Xie, Y.~Deng, D.~Yang, L.~Liang, J.~Ru, Y.~Yin, and Y.~Zou.
\newblock Vargpt-v1. 1: Improve visual autoregressive large unified model via iterative instruction tuning and reinforcement learning.
\newblock {\em arXiv preprint arXiv:2504.02949}, 2025.

\end{thebibliography}
}

\end{document}